\documentclass[draft]{agujournal2019}
\usepackage{url} 
\usepackage{soul}
\usepackage{float}
\usepackage{caption}
\usepackage[finalnew]{trackchanges} 
\usepackage{soul}
\draftfalse

\journalname{Earth and Space Science}

\begin{document}

%
%


\title{Fusing Multi- and Hyperspectral Satellite Data for Harmful Algal Bloom Monitoring with Self-Supervised and Hierarchical Deep Learning}

%
%




\authors{Nicholas J. LaHaye\affil{1,2}, Kelly M. Luis\affil{2}, and Michelle M. Gierach\affil{2}}

\affiliation{1}{Spatial Informatics Group, Pleasanton, CA, USA}
\affiliation{2}{Jet Propulsion Laboratory, California Institute of Technology, Pasadena, CA, USA}






\correspondingauthor{Nicholas LaHaye}{nlahaye@sig-gis.com}



\begin{keypoints}
\item This paper demonstrates the successful application of self-supervised learning (SSL) for U.S. coastline HAB monitoring.
\item The described SSL approach enables single and multi-sensor ocean color observations of HAB events.
\item Initial testing with new hyperspectral instruments demonstrates the potential to meet NASA program-of-record needs.
\end{keypoints}

%
%

%
%


\begin{abstract}
We present a self-supervised machine learning framework for detecting and mapping the severity and speciation of harmful algal blooms (HABs) using multi-sensor satellite data. By fusing reflectance data from operational \add[]{polar-orbiting satellite-based} instruments (VIIRS, MODIS, \add[]{OLCI, and OCI}) with TROPOMI solar-induced fluorescence (SIF), our framework, called SIT-FUSE, generates HAB severity and speciation products without requiring per-instrument labeled datasets. The framework employs self-supervised representation learning and hierarchical deep clustering to segment phytoplankton \change[]{concentrations}{cell abundance} and species into interpretable classes, validated against in-situ data from the Gulf of Mexico and Southern California (2018–2025). Results show strong agreement with total phytoplankton, \change[]{Karenia brevis}{\textit{Karena brevis}} \remove[]{, Alexandrium spp., } and \change[]{Pseudo-nitzschia spp.}{\textit{Pseudo-nitzschia spp.}} measurements. This work advances scalable HAB monitoring in \change[]{label-scarce environments}{environments where ground truth observations are limited}, while enabling exploratory analysis via hierarchical embeddings - a critical step toward operationalizing self-supervised learning for global aquatic biogeochemistry.
\end{abstract}

\section*{Plain Language Summary}
Harmful algal blooms (HABs) \change[]{are large outbreaks of algae}{are increases in the abundance of algae} in oceans and lakes that can harm people, animals, and the environment. While monitoring these blooms is essential for protecting public health, fisheries, and coastal economies, blooms can appear quickly and change rapidly. Traditional \add[]{satellite-based monitoring methods} \remove[]{rely on satellites and }require many matchups with ground-based measurements and either custom algorithms for each instrument or algorithm parameter updates for each new instrument, making them time-consuming and expensive to develop.
This paper introduces a new approach for monitoring, using a flexible software framework called \add[]{Segmentation, Instance Tracking, and data Fusion Using multi-SEnsor imagery (SIT-FUSE)}. SIT-FUSE uses artificial intelligence to automatically detect, track, and map HABs by combining information from multiple types of satellite sensors, including both standard and state-of-the-art instruments. Unlike previous methods, SIT-FUSE can work with little to no pre-existing labeled data, while centering the subject matter experts in the modeling and validation processes. It can also merge data from different satellites to create more detailed and frequent maps of a bloom’s evolution and its severity, and even distinguish between phytoplankton species.
The system successfully identified and mapped HABs, including specific harmful species, and performed well even in complex coastal waters. By using SIT-FUSE, scientists and managers can obtain accurate, timely information on HABs for societal and ecosystem benefits.

%
%

\section{Introduction}
\label{sec1}
Phytoplankton, microscopic photosynthetic algae, are the base of the marine food web. When certain phytoplankton species are in high \change[]{concentrations}{cell abundance}, they can cause severe environmental, human health, and economic problems. Harmful algal blooms (HABs) are most often associated with events where toxin-producing phytoplankton bioaccumulate throughout the marine food web. The propagation of toxins leads to fish kills, marine mammal and shellfish mortality, closures of fisheries and tourism operations, and even increased human hospitalizations related to toxin ingestion or airborne exposure \cite{anderson2001phytoplankton}. The impacts are estimated to cost the U.S. \$10-100 million annually \cite{anderson2000estimated, Hoagland2002}. With frequency, severity, and geographic distribution of global HABs projected to expand with climate change \cite{Gobler2020}, early and real-time detection of bloom events is a priority for decision-making. 

Multiple remote sensing platforms have been leveraged for gaining real-time information for monitoring and management. \change[]{Recent work has shown that remote sensing can reduce annual potential HAB-associated costs on the order of \$5.7- 316 million dollars}{Recent work has shown that remote sensing can reduce annual damages from HABs on the order of \$158-\$234 million dollars} \cite{geoxo}. Red tide events, associated with \change[]{Karenia brevis}{\textit{Karenia brevis}}, are routinely monitored along the West Florida Shelf with multispectral ocean color remote sensing in the visible to near-infrared spectrum. Daily multispectral ocean color products include normalized fluorescence line height (nFLH) and chlorophyll-a (chl-a). The advent of spaceborne hyperspectral or spectroscopic remote sensing instruments, such as \add[]{the Plankton, Aerosol, Cloud, ocean Ecosystem (PACE) mission's Ocean Color Instrument (OCI), PRecursore IperSpettrale della Missione Applicativa (PRISMA), and Earth Surface Mineral Dust Source Investigation (EMIT), which is on the International Space Station (ISS)}, provides critical spectral information that enables identification of phytoplankton community composition, including HABs. However, these optical observations are limited to clear-sky days, and complex water types can complicate the accurate retrieval of these products. \add[]{Gap-filling methods such as Dynamincal Interpolation Empirical Orthogonal Function (DINEOF) have been operationalized to increase the frequency of retrievals. However, many of these statistical methods rely on linear interpolation and do not incorporate geospatial context.} On the other hand, recent advancements in red solar-induced chlorophyll fluorescence (SIF) measurements from Sentinel 5P/TROPOMI, as demonstrated by \cite{Khler2020, Luis2023, Madani2024}, can retrieve phytoplankton fluorescence information under optically variable atmospheric and water-column conditions, but these methods are generally at coarser spatial resolutions (7 km).

Traditionally, detecting and monitoring environmental events such as phytoplankton blooms and harmful algal blooms (HABs) \change[]{with a single remote sensing instrument has relied on highly specialized retrieval algorithms tailored to each sensor}{has relied on the development of highly specialized retrieval algorithms tailored to specific band configurations}. This process is resource-intensive, often requiring detailed domain expertise for parameter selection and instrument-specific calibration, and significant manual effort to consistently track features across multiple scenes. Recently, the advent of supervised deep learning methods (e.g., Convolutional Neural Networks, or CNNs) has accelerated the development of new, more automated retrieval algorithms for these tasks, offering the potential to streamline and improve environmental monitoring efforts \cite{Kravitz2024}, but requiring large pre-existing label sets to achieve accurate results. 

In previous work, we showed that a self-supervised Deep Belief Network (DBN) trained on \change{L1 (instrument reflectance) or L2 (instrument radiance}{L1 (instrument radiance) or L2 (instrument reflectance)} data can segment geophysical objects when combined with unsupervised clustering \cite{LaHaye2019}. This approach offers two key advantages: (1) Resolution and Instrument Flexibility - The method adapts to diverse spatial, spectral, and temporal resolutions, enabling cross-instrument object detection and tracking; (2) Label Efficiency - Instead of labor-intensive per-instrument labeling, it applies coarse context assignments post-segmentation to a limited set of training scenes, making it viable for \change[]{label-scarce scenarios}{scenarios where ground truth labeling is limited}. Follow-up studies confirmed these foundational concepts by implementing and testing a streamlined architecture for both atmospheric and land-surface classification across heterogeneous inputs, including varying spectral, spatial, temporal, and multi-angle remote sensing data \cite{LaHaye2021}. Building on this foundation, we have shifted from unsupervised clustering to self-supervised deep clustering (detailed in the Methods section). This evolution allows the framework to leverage vast, unlabeled training data from diverse scenes and move beyond the constraints of pre-existing labeled datasets required by traditional supervised methods. The fully self-supervised paradigm enhances scalability and generalizability, particularly for applications where substantial manual labeling is impractical.

In our latest work \cite{LaHaye2025a, LaHaye2025b}, we expanded our machine learning framework into the SIT-FUSE (Segmentation, Instance Tracking, and data Fusion Using multi-SEnsor imagery) library, an open-source system for segmenting, tracking, and analyzing geophysical objects in remote sensing data from multiple platforms and modalities. The framework supports diverse encoder architectures, including Deep Belief Networks (DBNs) (convolutional and standard), Vision Transformers \add[]{(ViTs)}, \add{}{and} Convolutional Neural Networks (CNNs). This evolution also replaces traditional unsupervised clustering with deep-learning-based clustering, enhancing adaptability, reproducibility, and precision. This approach offers several distinct advantages, making it especially powerful and flexible for earth observation. SIT-FUSE is not restricted to any particular remote sensing platform or to fixed spatial or spectral resolutions, making it broadly applicable and compatible with data from a wide range of sensor types. This flexibility enables researchers to integrate and analyze observations collected across diverse instruments without platform-specific constraints. The method is also capable of recognizing and tracking geophysical features as they appear in data from multiple instruments, bringing together separate streams of information and integrating them within self-supervised model architectures. Importantly, this kind of data fusion enables analysis across a diverse set of scenes and problem types—even when there are few or no labeled examples. Unlike strictly supervised machine learning methods, it can meaningfully learn from large pools of unlabeled data, making it invaluable for advancing research in domains where annotated datasets are scarce.

Here, we demonstrate SIT-FUSE’s versatility in addressing diverse environmental challenges beyond the original validation datasets and its original application of wildfire. This study implements a self-supervised machine learning framework to address the automated detection and mapping of phytoplankton bloom \change[]{concentrations}{cell abundance} and speciation, with an emphasis on harmful algal blooms (HABs). The analysis utilizes temporal sequences of surface reflectance data obtained from multiple multispectral remote sensing platforms, enabling robust feature identification and classification across heterogeneous datasets. The foci of this study is sequences of \add[]{fully atmospherically corrected} surface reflectance data \add[]{(using each mission's operational atmospheric correction pipeline, via the remote sensing reflactance (RRS) product)} acquired by multiple multispectral remote sensing instruments from 2018-2019, and then a smaller test case for a hyperspectral instrument, NASA’s PACE-OCI, in 2024-2025.

\section{Materials and Methods}
\label{sec2}
\subsection{Study Areas}
\label{subsec21}
\subsubsection{Southern California}
\label{subsubsec211}

Southern California waters host several HAB \change[]{species}{taxa of concern} \add[]{.} \change[]{Pseudo-nitzschia species (P. spp.)}{\textit{Pseudo-nitzschia species (P. spp.)}} produce the neurotoxin domoic acid, \add[]{which is }responsible for amnesic shellfish poisoning and mass strandings of marine mammals and seabirds \cite{Bates2018, Smith2018} \remove[]{while Alexandrium species (A. spp.) generate saxitoxins that cause paralytic shellfish poisoning}. \add[]{While multiple \textit{P. spp.} are present in this region, \textit{P. delicatissima} and \textit{P. seriata} are included in this analysis and differ markedly in toxic potential. \textit{P. seriata} is more frequently associated with elevated domoic acid production and harmful impacts, whereas \textit{P. delicatissima} is typically less toxic.}

\change[]{Collectively, }{\textit{P. spp.}} has been implicated in large-scale fish kills, shellfish harvesting closures, and ecosystem disruptions \cite{Lewitus2012, McCabe2016}. While blooms of these taxa occur throughout the California Current System, they are \change[]{persistent}{especially frequent} and impactful \change[]{along}{in} the Southern California Bight, where coastal topography, nutrient dynamics, and circulation features converge to favor HAB development. \add[]{In this region, statistical relationships between satellite ocean color observations and in situ phytoplankton counts, domoic acid particulates have been integrated into hydrodynamic models to map \textit{Pseudo-nitzchia} HAB events} \cite{seegers2015, anderson2009, anderson2011}.

The California Current, flowing equatorward along the coast, interacts with semi-enclosed embayments, coastal headlands, and islands to create retention zones that promote HAB accumulation \cite{Horner1997, Smith2018}. Seasonal upwelling delivers nutrient-rich waters to the surface, fueling phytoplankton growth, while \add[]{subsequent} relaxation of upwelling and \change[]{subsequent}{water-column} stratification can favor the dominance of toxigenic species. \change[]{P. spp.}{\textit{P. spp.}} blooms commonly develop in spring and summer, coinciding with strong upwelling and nutrient availability, but they can also recur in fall during stratified, warmer conditions \cite{Smith2018}. \remove[]{\textit{A. spp.} blooms are typically less predictable but often appear in late spring to summer, when favorable currents and water column structure promote population growth and accumulation.} The complex bathymetry of the Southern California Bight with broad shelves, submarine canyons, and island wakes further enhances retention and transport, allowing blooms to intensify and persist near shore \cite{Pitcher2010, Hickey2008}.

\subsubsection{Gulf of Mexico}
\label{subsubsec212}
\change[]{Karenia brevis}{\textit{Karenia brevis}} is by far the most frequent and consequential HAB in the Gulf of Mexico. This dinoflagellate drives the region’s well-known “red tides,” releasing brevetoxins that cause widespread fish kills, shellfish toxicity, and respiratory irritation when aerosolized near shore \cite{Steidinger2009}. While \change[]{K. brevis can be detected across the Gulf, blooms occur most reliably along the West Florida Shelf (WFS), where nearly annual events unfold}{\textit{K. brevis} is present across the Gulf of Mexico, with blooms occurring most consistently along the West Florida Shelf, where they can be observed and tracked using multispectral visible–near-infrared ocean color remote sensing} \cite{Cannizzaro2009, Tomlinson2009, Stumpf2003, Carvalho2011, HU2005}.

The Gulf of Mexico’s physical setting is central to shaping \change[]{K. brevis}{\textit{K. brevis}} dynamics. As a semi-enclosed basin, it is dominated by the Loop Current, which flows northward through the Yucatán Channel before exiting via the Florida Straits \cite{He2003, Weisberg2017}. On the WFS, a broad, gently sloping continental margin provides extensive shallow habitat that interacts strongly with this circulation\add[]{.} When the Loop Current or associated eddies impinge on the shelf slope, they can trigger upwelling and entrain nutrient-rich waters onto the shelf \cite{Walsh2003, Weisberg2022}. Combined with wind forcing and Ekman transport, these processes help sustain offshore populations and carry them shoreward. The geometry of the WFS, including its wide, shallow expanse and limited cross-shelf exchange, further promotes accumulation. At the same time, mesoscale eddies, fluid transport barriers, and bathymetric features concentrate blooms rather than dispersing them \cite{Weisberg2022}. \change[]{K. brevis}{\textit{K. brevis}} blooms also follow a pronounced seasonal cycle. On the WFS, they typically initiate offshore in late summer, when warm, stratified waters favor growth. Through the fall, circulation patterns and winds transport populations toward the coast, where they often reach peak intensity in autumn and early winter (August–December, sometimes extending into January) \cite{Walsh2006, Weisberg2019}.

\subsection{Data}
\label{subsec22}
\subsubsection{Input Datasets}
\label{subsubsec221}
For the initial tests, data from \add[]{orbital instruments including the Visible Infrared Imaging Radiometer Suite (VIIRS) instruments onboard the Suomi National Polar‑orbiting Partnership (SNPP) and Joint Polar Satellite System‑1 (JPSS1) platforms, the Moderate Resolution Imaging Spectroradiometer (MODIS) instrument onboard the Aqua platform, and the Ocean and Land Color Instrument (OLCI) onboard the Sentinel-3A, and Sentinel-3B platforms} were used in the time period of June 1, 2018 to December 31, 2019 to overlap with testing and analysis done in \cite{Luis2023}. The period from June 1, 2018, to August 31, 2019, was used for training, and the period from September 1, 2019, to December 31, 2019, for testing. For the PACE tests, the time range from March 5, 2024, to March 31, 2025, was used. The period from March 5, 2024, to January 31, 2025, was used for training, and the period from February 1, 2025, to March 31, 2025, was used for testing. Surface reflectance was chosen here for two reasons: 1) the likelihood of missed latent patterns, especially in complex waters, as mentioned above, when only using downstream ocean color parameters to proxy phytoplankton presence, and 2) the inconsistent availability of various parameters across all of the instruments used. 

The Sentinel-5P \add[]{TROPOspheric Monitoring Instrument (TROPOMI)-based} red SIF (TROPOSIF) products have been generated based on the retrieval approach from \cite{Khler2020}, and the data \change[]{is}{are} hosted on \url{ftp://fluo.gps.caltech.edu/} or data.caltech.edu. \cite{Khler2020} implemented a variant of an established far-red SIF retrieval scheme \cite{Joiner2013, Guanter2012, Damm2015, Kohler2015, Khler2018} to estimate red SIF from TROPOMI measurements for aquatic science. The TROPOSIF data generated for the 2018-2019 time period has only been produced in a daily ungridded format, so this data was taken and gridded at its native 7km resolution. \cite{Khler2020, Luis2023} highlighted the potential that red SIF has for improving our understanding of global phytoplankton photosynthesis and HABs. Specifically, \cite{Luis2023} found that red SIF provided more than twice as much data as nFLH and thus provided a new monitoring capability to obtain critical information on HABs. The TROPOSIF data is not available for the 2024-2025 time period, so it was not used in conjunction with the PACE data here. The test areas were the Gulf of Mexico and coastal Southern California.

\subsubsection{Dataset Preprocessing}
\label{subsubsec222}
For this task, the latest available versions of the 4km daily Level-3 Mapped (gridded) and fully atmospherically corrected surface reflectance data were used (Version 2 for VIIRS, Version 2022 for MODIS and Sentinel-3, and Version 3 for PACE). All data was re-projected to the WGS84 Latitude/Longitude projection at a 7km resolution to match the TROPOSIF product. All resampling and reprojections were performed using the open-source Python library pyresample. Also, as we are only looking at data over coastal and open-ocean regions, we applied the defined ocean basins mask from the open-source Python library regionmask, which was collected from Natural Earth. All data over land was masked and discarded for training and inference. \add[]{This focuses the model capacity on oceanic variability and avoids spurious clusters driven by land or mixed land–sea pixels, which are irrelevant for the task at hand.} For cases involving multi-sensor data, individual streams are colocated and concatenated along the channel dimension. The actual fusion is achieved during representation learning within each encoder, which integrates the combined inputs into unified feature representations. 

\begin{figure*}[h!]
\noindent\centering\includegraphics[width=0.7\textwidth, height=1.5in]{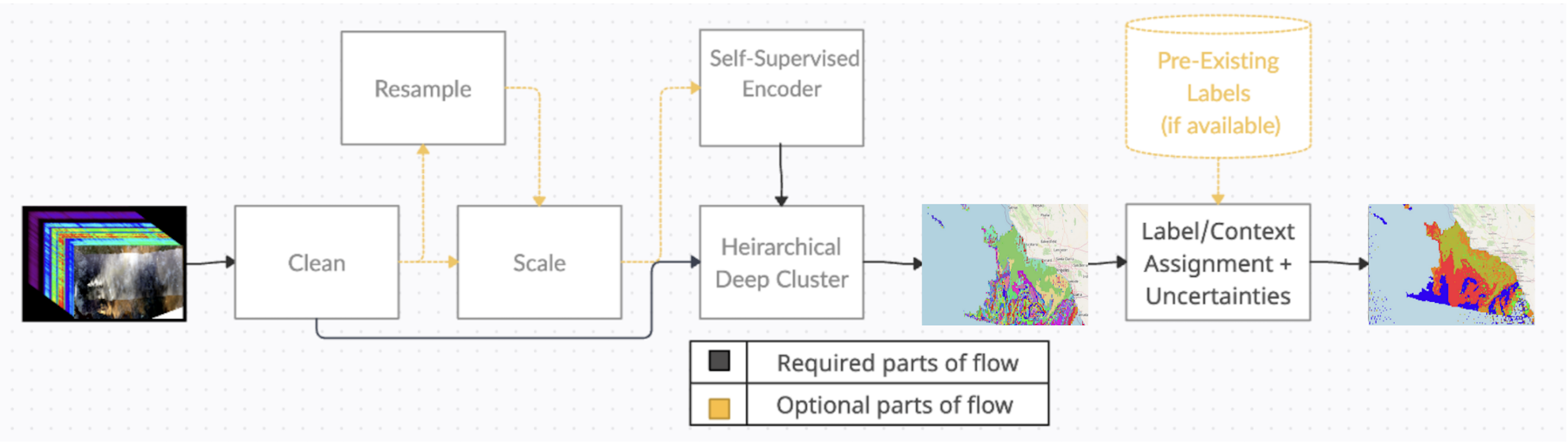}
\caption{A flowchart illustrating the processing pathway for a single input type, whether sourced from an individual instrument or a fused dataset, through SIT-FUSE.}\label{fig1}
\end{figure*}

Training samples were generated by extracting each pixel and its eight immediate neighbors across all spectral channels, forming a flattened vector to capture local spatial context. These vectors were standardized by subtracting per-channel means and scaling to unit variance, with statistics computed globally across the full training dataset. Before preprocessing, pixels containing fill values or data outside valid ranges \add[]{, as defined by input specification and metadata, } were systematically excluded. \add[]{For any sample, being a set of 9 pixels (center and immediate neighborhood), that is missing a pixel (for instance, given a pixel on the edge of an image, the neighborhood would be incomplete), or has an invalid/fill/masked value, we do not include this sample as a valid case for training or inference. Using small spatial neighborhoods preserves pixel-scale detail while still encoding geophysical structures such as fronts, filaments, and small coastal plumes, which are critical for HAB monitoring.} To ensure representative sampling of coastal conditions, approximately 3 million samples per encoder were subsampled from the training scenes using k-means clustering (50 classes) for stratification-a widely adopted, albeit naive, method for preserving spectral diversity \cite{MacQueen1967, Lin2022, Burke2023}. The 50-class threshold was determined via the elbow method \cite{Ashabi2020}. All visible spectral bands were used, and the same pixel subsets were used to train both the encoders and the deep clustering heads. Scenes used for training also provided the background for context assignments/in-situ matchups. While SIT-FUSE supports larger input tiles for convolutional DBNs, CNNs, and ViTs, non-convolutional DBNs achieve sufficient spatial context using the 3×3 pixel neighborhoods. The end-to-end data flow for each architecture is illustrated in Figure \ref{fig1}. \add[]{The general setup for the framework allows for a bypass of the encoder, passing data directly into the clustering head. Previous work on general validation of this approach demonstrated that as data complexity increases, the need for an encoder increases to ensure a robust representation from which interpretable information can be extracted. This can be thought of as the floor of the complexity scale when we consider encoder choice/complexity - (No Encoder, DBN, CNN, ViT, etc.). While this is the case for the framework generally, here we use an encoder for all datasets evaluated. Future work will examine and quantify tradeoffs in representational and final performance vs. the resource requirements of more complex, compute-intensive architectures and models.}

\subsection{Methods}
\label{subsec23}

\add[]{This section describes how SIT-FUSE turns multi-sensor satellite observations into HAB-relevant maps. We first explain how we learn generic ocean-color-related patterns from the observations (via self-supervised encoders), then how we group similar pixels into various regimes (via deep clustering), and finally how we attach ecological meaning to a subset of those groups using HAB-focused in situ data (via context assignment). Throughout, our design choices are guided by constraints typical of remote sensing for HAB monitoring: limited in situ matchups, heterogeneous sensors, and the need for interpretable, operationally scalable products.}

\subsubsection{Self-Supervised Representation Learning}
\label{subsubsec231}

SIT-FUSE is architected as a modular platform that accommodates a wide range of encoder and earth observation foundation models, each leveraging self-supervised representation learning strategies. Supported encoder types include Deep Belief Networks (DBNs) trained via contrastive divergence, Convolutional Neural Networks (CNNs) enhanced with residual connections and pixel-wise contrastive optimization, and transformer-based architectures such as the Image-Joint Embedding Predictive Architecture (I-JEPA) and Masked Autoencoding (MAE). In addition, the framework integrates pre-trained earth observation foundation models like Clay and Prithvi \cite{Grill2020, He2022, Liao2022, Assran2023}). Experimental implementations here prioritized 2–3-layer DBNs due to their parameter efficiency ($\sim2$ million parameters) compared to larger architectures (100 million–10 billion parameters), while maintaining competitive representational capacity \cite{Liao2022}. \change[]{Validation across single-instrument and multi-sensor fusion datasets confirmed DBNs’ structural interpretability, downstream task performance, and computational sustainability, critical for operational scalability.}{For this application and case study, we prioritize DBNs because they are compact enough to run routinely on large, multi-year datasets while still capturing the main spectral–spatial patterns that distinguish bloom and non-bloom waters.} While larger models dominate recent literature, as deep learning approaches gain more operational adoption and visibility, practitioners must continue to consider and evaluate smaller and potentially more efficient architectures along with the much larger and more novel architectures, to keep energy and compute resource consumption as low as possible, while still increasing adoption of these techniques to balance accuracy with energy/compute constraints-a priority for global operational deployment. Ongoing work quantifies segmentation performance and geographic coverage tradeoffs across encoder complexities, with findings to be detailed in subsequent publications.

The DBN architectures used here employed architecture-driven feature expansion, projecting pixel neighborhoods into higher-dimensional latent spaces to capture nonlinear patterns more effectively than lower-dimensional kernelization approaches \cite{Harrington2018, Harrington2020, Sobczak2019}. Encoder depth and hidden/output parameters were dynamically adjusted based on input spectral resolution and associated latent pattern complexity. \add[]{Intuitively, the encoder stage creates a fingerprint for each pixel neighborhood, summarizing its local spectral shape and context in a way that downstream clustering can exploit without needing explicit chlorophyll or other phytoplankton-related retrievals. The outcome of this step is a set of trained encoders that maps each pixel (with local spatial context) into an embedding space with no semantic or physical meaning yet. At this stage, no species, cell abundance, or HAB information has been inferred. Only latent representations have been learned.}

\subsubsection{Deep Clustering}
\label{subsubsec232}

To generate context-free segmentation maps from per-pixel embeddings, we employ Invariant Information Clustering (IIC), a deep learning approach that replaces traditional agglomerative methods (e.g., \add[]{Balanced Iterative Reducing and Clustering using Hierarchies (BIRCH)}) in our framework. \change[]{This transition addresses critical limitations in computational efficiency: neural network-based clustering implemented via PyTorch reduces training/inference times, memory overhead, and model portability compared to conventional scikit-learn workflows. }{Using a neural-network-based clustering implementation mainly improves runtime and portability; the key scientific idea is that we preserve the ability to resolve multiple, nested habitat and bloom regimes rather than a single flat classification.} The IIC-based method optimizes cluster assignments by maximizing the mutual information between an input sample, x, and its perturbed counterpart, $x'$ \cite{Ji2018}. For our purposes, perturbations are introduced as Gaussian noise applied to encoder outputs. 

\begin{figure}[h!]
\noindent\centering\includegraphics[width=2.5in]{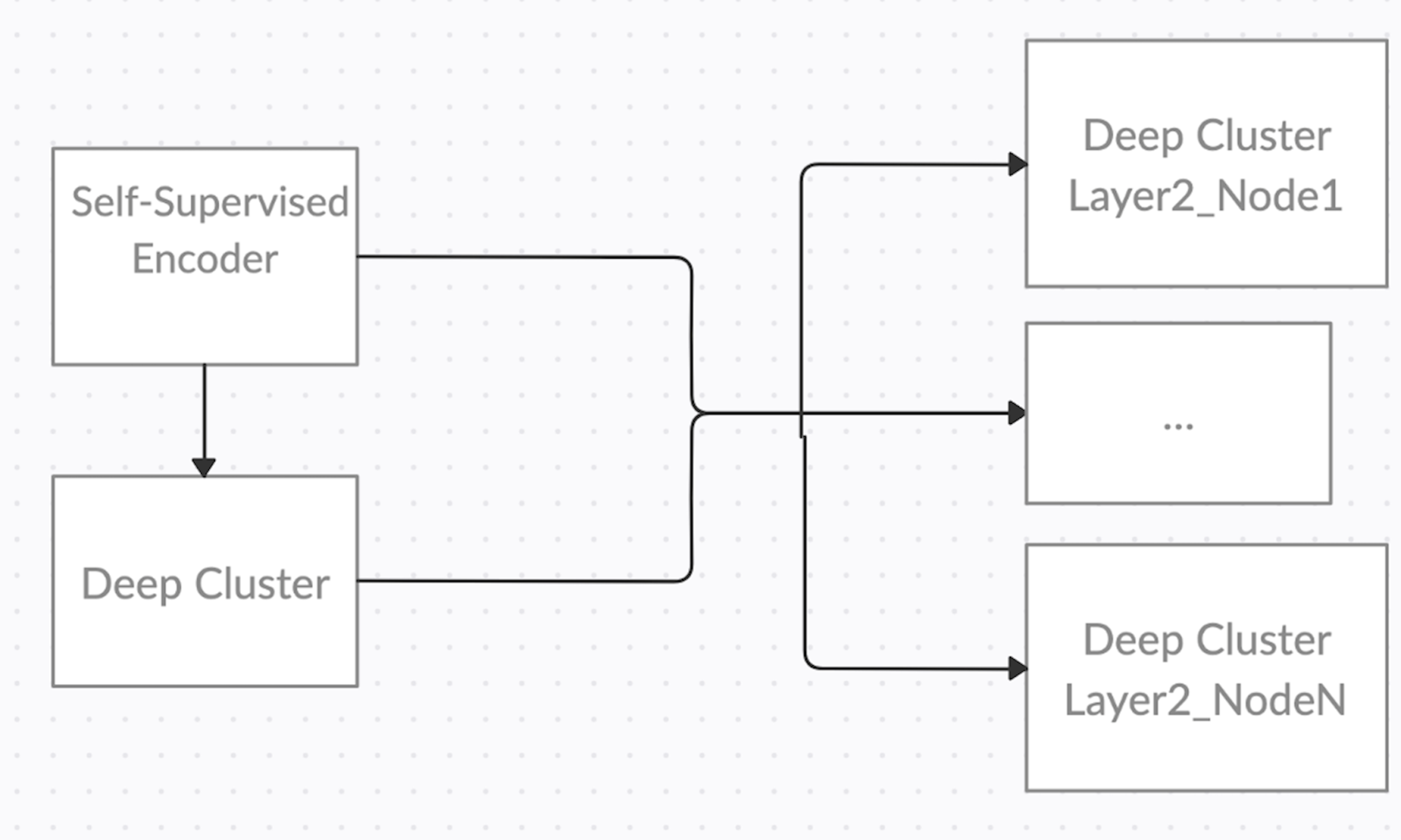}
\caption{A two-layer hierarchical deep clustering architecture is illustrated, where each 'Cluster' box consists of fully connected neural network layers that interface with the encoder and are trained using the IIC loss (Invariant Information Clustering) function. In this structure, each child node is specifically trained and makes predictions only for the subset of data assigned by its parent cluster. This hierarchical setup enables the deep clustering framework to progressively refine and organize data into multiple, nested levels, facilitating detailed exploration and characterization across varying degrees of specificity.}\label{fig2}
\end{figure}

To emulate the multi-tiered representative capabilities of agglomerative clustering, we designed a tree-structured hierarchical clustering system (Figure \ref{fig2}). The root node partitions data into an initial set of coarse classes (here, 800), while child nodes refine these into separate sets of subclasses (here, 100 each), trained exclusively on samples inherited from their parent clusters. \add[]{At the coarsest level, for this application, this tends to separate broad water types (e.g., low-biomass open ocean vs productive, optically complex coastal waters); the child nodes then resolve finer distinctions such as high-biomass HAB regimes within coastal waters or different phytoplankton assemblages within the same region. Beyond segmentation accuracy, this structure facilitates exploratory data analysis by revealing latent connections between classes and class hierarchies. The resulting clusters are context-free statistical groupings that do not yet correspond to species, cell abundance levels, or physical oceanographic variables.} This top-down hierarchy enables scene segmentation at user-defined levels of specificity, with class relationships explicitly encoded in the tree's topology. \remove[]{To our knowledge, this represents the first implementation of IIC/deep clustering for segmentation in a hierarchical configuration.} While current implementations require manual specification of hierarchy depth (two levels in this study), future iterations could integrate automated node splitting. \add[]{The output of this step is not 'HAB' vs 'no HAB' yet, but a set of objectively derived ocean surface and bloom-pattern classes that can later be mapped to ecologically meaningful categories using in situ observations.}

\subsubsection{Context Assignment}
\label{subsubsec233}

\remove[]{To assign the desired context to the context-free segmentation products, we used in situ data collected around coastal Florida and Southern California. For Southern California, this included data from the California Harmful Algal Bloom Monitoring and Alert Program}\remove[]{, and for Florida, this included data from the Florida Fish and Wildlife Conservation Commission's Recent HAB Events dataset.}\add[]{To assign the desired context to the context-free segmentation products, we used in situ data collected from coastal monitoring programs in Florida and Southern California. In Southern California, in situ observations were obtained from the California Harmful Algal Bloom Monitoring and Alert Program }(\url{https://calhabmap.org/} ; CalHABMAP)\add[]{, which collects weekly phytoplankton and water quality measurements at nine long-term monitoring locations along the California coast. In Florida, in situ data were drawn from the Florida Fish and Wildlife Conservation Commission’s (FWC) Recent Harmful Algal Bloom Events dataset }(\url{https://bit.ly/3LGAwZx})\add[]{, which is supported by the Fish and Wildlife Research Institute’s (FWRI) HAB monitoring program. The FWRI HAB group monitors more than 100 locations statewide, with sampling frequencies ranging from weekly to monthly, depending on bloom conditions, resource availability, and management priorities. Differences in spatial coverage and sampling frequency between these programs reflect regional monitoring objectives and logistical constraints and are explicitly accounted for in the context assignment process. While the Florida program provides broader spatial coverage with variable temporal resolution, the Southern California program offers higher temporal consistency at fewer sites. These characteristics inform the interpretation of the resulting context-labeled products and help constrain satellite-derived patterns within the observational limits of each monitoring system.}

\begin{figure*}[p]
\begin{minipage}{\textwidth}
\noindent\centering\includegraphics[width=\textwidth]{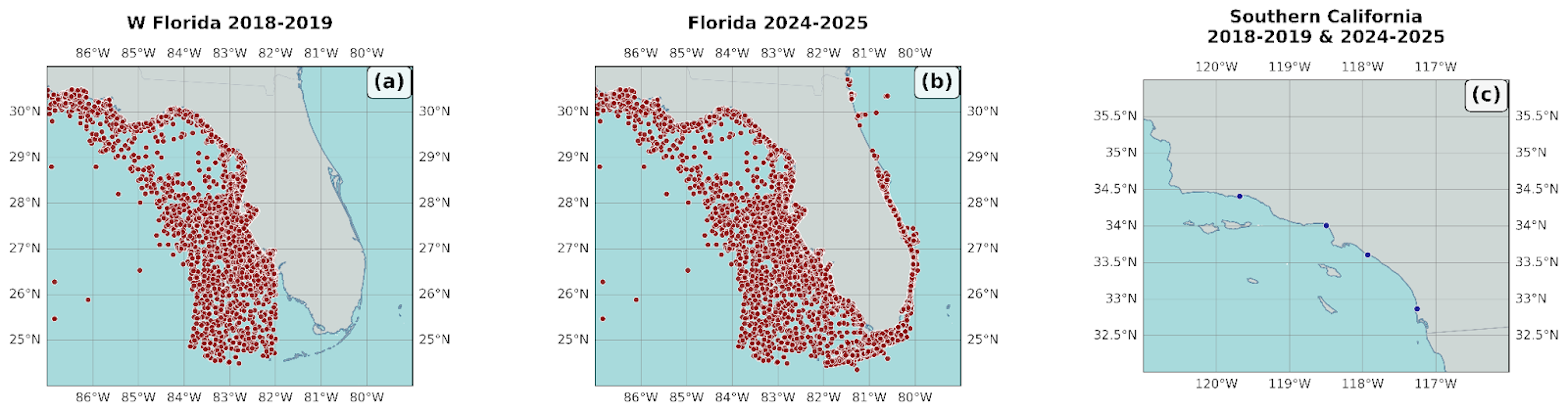}
\caption{A depiction of all of the locations where in situ data was collected and used for context assignment and validation:  a) West Florida cases in 2018-2019, b) Florida 2024-2025, and c) Southern CA in the 2018-2019 and 2024-2025 cases. The actual process of context assignment is done by generating simple histograms, or counts of overlap between a specific index within the binned set of phytoplankton \change[]{concentrations}{cell abundance} and each label in the context-free segmentation products. The final assignment is to assign a context-free label to the phytoplankton \change[]{concentrations}{cell abundance} bin with which it most frequently overlaps. More sophisticated thresholding and overall assignment techniques can be applied, but we found this simple approach to be suitable for the cases tested.}\label{fig3}
 \end{minipage}
    \hfill
     \vspace{1cm}
  \begin{minipage}{\textwidth}
\noindent\centering\includegraphics[width=\textwidth]{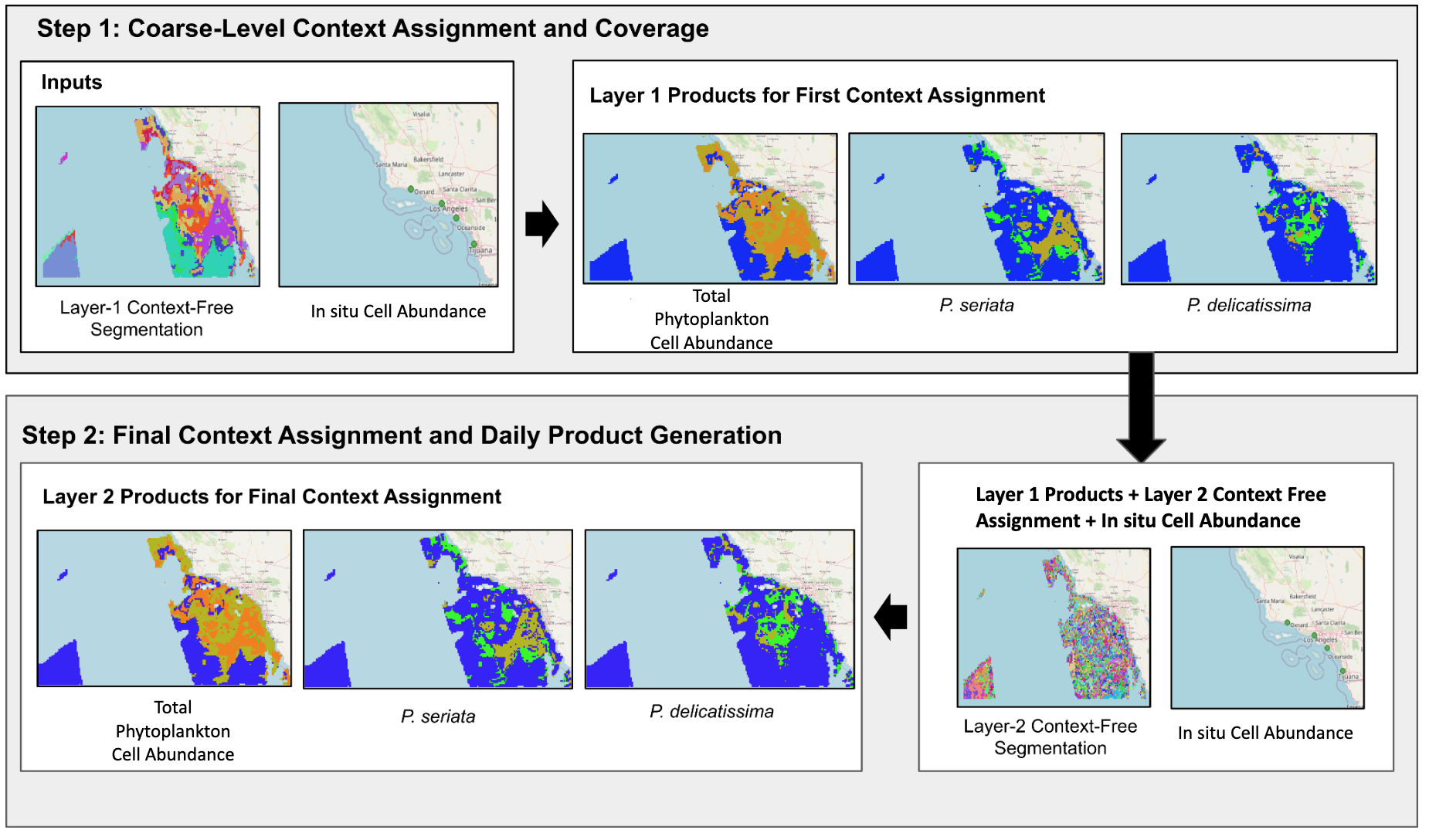}
\caption{A depiction of the multi-tiered context assignment process based on the hierarchical context-free segmentation products. Step 1 (top) consists of finding the context-free labels that best match with different binned phytoplankton or speciated HAB \change[]{concentrations}{cell abundance} levels. Once this is done, the first context assignment is done accordingly. Next, Step 2 (bottom) consists of applying the same process to the Layer-2 context-free segmentation product, and then supplementing the agreement computations there by also looking at agreement between the Layer-2 context-free labels and the \change[]{concentrations}{cell abundance} labels assigned in step 1 over the scene. This process is performed collectively across the training set of scenes, and Step 2 provides the final context assignment to be used for all scenes. \add[]{This two-step, hierarchical assignment lets us first capture where phytoplankton are present in a broad sense, and then refine within those areas to distinguish specific HAB species and severity levels where enough in situ coverage exists. The colors within the context-free segmentation plots depict various context-free classes, while the colors in the speciated plots represent cell abundance levels of different phytoplanton species/populations. Further information will be provided below on the color mapping of cell abundance; this figure is only intended to depict the process.}}\label{fig4}
 \end{minipage}
\end{figure*}

Figure \ref{fig3} depicts the locations of the in situ data utilized in this study. Over the entire training set, we removed all samples with a depth greater than 1m. We identified a minimum radius within which a large enough set of pixel-in-situ observation matchups exists over areas with valid data. Because there is a much larger in-situ network off the coasts of Florida, and many sites are farther offshore, we were able to use a smaller radius for the Florida test cases: 0.0225 decimal degrees, or ~2.5 kilometers. Because of the coarseness of the coastal masking and the limited in-situ sites in Southern California connected to piers, a larger radius of 0.09 decimal degrees (~10km) had to be used. As shown in the results, we were able to generate representative products in both regions, but a larger, more widely distributed set of in situ sites is always preferred.

For the Florida sites, the in situ data provides measurements of \change[]{K. brevis}{\textit{K. brevis}} \change[]{concentrations}{cell abundance}, and for the Southern California sites, the data includes a total phytoplankton \change[]{concentrations}{cell abundance}, as well as separate \change[]{concentrations}{cell abundance} of 12 different species. For this study, in the Southern California cases, we focused on \change[]{Pseudo-nitzschia delicatissima}{\textit{Pseudo-nitzschia delicatissima}} and \change[]{Pseudo-nitzschia seriata}{\textit{Pseudo-nitzschia seriata}} \remove[]{, and \textit{Alexandrium spp.}}, as well as the total phytoplankton value. \add[]{We acknowledge that \textit{P. delicatissima} and \textit{P. seriata} differ in their toxicological significance, with \textit{P. seriata} more frequently associated with domoic acid production and harmful algal bloom events. Accordingly, detection of \textit{P. delicatissima} in this study is not intended to imply equivalent HAB risk, but rather to illustrate the framework’s ability to resolve and map multiple phytoplankton taxa with distinct ecological and management implications.} Also, although there is a hierarchical relationship between total phytoplankton and the \change[]{concentrations}{cell abundance} of each species, and there are likely other correlations between the \change[]{concentrations}{cell abundance} of each of the species, the context assignment for each specific \change[]{concentrations}{cell abundance} map, while derived from the same context-free segmentation data, is done separately to ensure we maximize coverage for each separate class. This approach has also proven successful for other SIT-FUSE applications, such as fire and smoke segmentation.

\add[]{While the abundance binning thresholds can be updated by region and species within the framework, for consistency across the manuscript, we applied a common legend and categorical structure when visualizing phytoplankton species. For \textit{K. brevis}, well-established cell abundance thresholds and severity categories are defined and routinely used by management authorities such as the Florida Fish and Wildlife Conservation Commission }\newline(\url{https://myfwc.com/research/redtide/statewide/})\add[]{, and these standards were adopted directly in this study. In contrast, for \textit{Pseudo-nitzschia spp.}, we did not identify a similarly standardized, multi-tiered classification framework beyond the commonly cited threshold of} $> 10000$ \add[]{cells/L associated with bloom or toxic conditions. As a result, and to maintain consistency between figures and text, we applied the \textit{K. brevis} categorical framework uniformly across species, with the understanding that the label “Low” in the \textit{P. spp} maps reflects this shared visualization convention rather than a statement about toxicity risk. We have clarified this distinction to emphasize that \textit{P. spp} toxicity can occur at cell abundances classified as “Low” under this generalized framework, and that these categories should be interpreted as relative cell abundance classes rather than toxicity thresholds.}

Because we produce two layers of hierarchical context-free segmentation data, we can use them together to drive coarse- and fine-scale context assignment. To do this, we first do the context application process with the coarser, layer-1 context-free segmentation. This provides us with an initial mapping and \change[]{broader coverage of the areas containing phytoplankton.}{broader coverage of the regions containing phytoplankton of interest for each domain, constrained and informed by available in situ observations.} Then, we both directly apply the context assignment to the layer-2 context-free segmentation in the same way we did with the Layer-1 products, and supplement by doing the same context assignment process between the Layer-2 context-free segmentation and the phytoplankton \change[]{concentrations}{cell abundance} map produced from Layer-1. There is typically overlap and agreement, but running the process in a tiered way does improve specificity, \add[]{relative to the in situ observations,} as can be seen by the speciated \change[]{concentrations}{cell abundance} changes from the Layer-1 products to the Layer-2 products in Figure \ref{fig4}. \add[]{After the initial context assignment process is done, a single lookup table is generated from the process described above, and is applied to all subsequent scenes passing through the system.}

\clearpage
\begin{figure*}[t]
\noindent\centering\includegraphics[width=\textwidth]{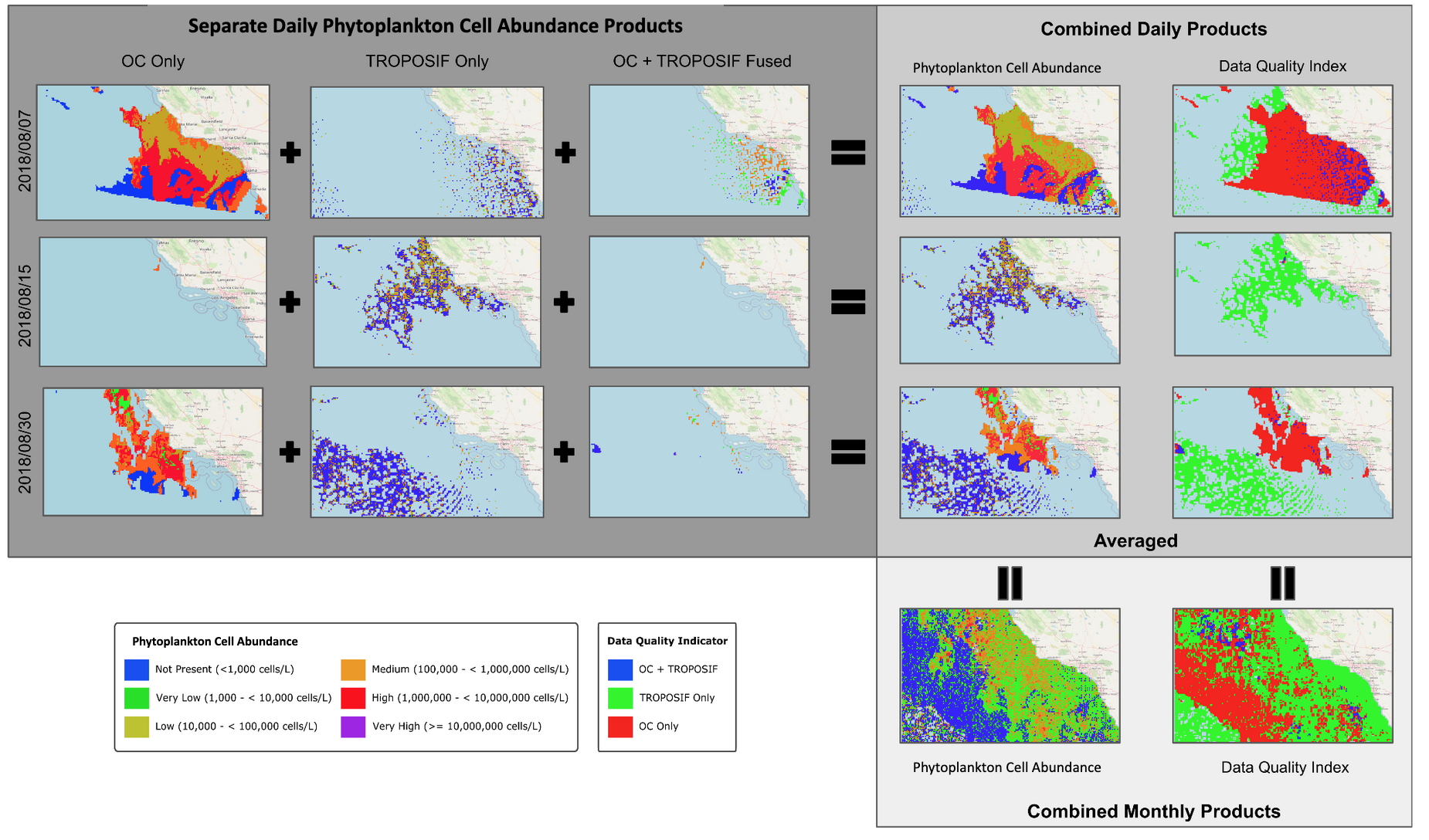}
\caption{A depiction of the combination of the various data streams. First ocean color (OC only, TROPOSIF only, and TROPOSIF + OC \change[]{concentrations}{cell abundance} datasets are combined into a single phytoplankton or speciated HAB \change[]{concentrations}{cell abundance} product. An associated Data Quality Indicator (DQI) product is also generated, indicating which data stream a given pixel originated from.  Lastly, monthly averages were generated for both \change[]{concentrations}{cell abundance} and DQI.}\label{fig5}
\end{figure*}

\subsubsection{Combining Data Streams}
\label{subsubsec234}
Because we want to maximize the coverage of each product, where available, we generate separate outputs for each ocean color (OC) instrument (VIIRS, MODIS, S3, PACE), TROPOSIF, and OC instruments + TROPOSIF. Each output stream has its own set of context-free segmentation labels and, therefore, its own context assignment. \add[]{This design mirrors how operational centers already handle multi-sensor products: each instrument contributes where it has coverage, and fused maps expose where estimates come from, which is essential for interpreting potential biases.} Once context has been assigned, each data stream has a phytoplankton (and separately speciated) \change[]{concentrations}{cell abundance} product. These products are then merged on a per-instrument-set / per-day basis. Figure \ref{fig5} depicts the data stream combination process.

For now, we kept data streams for each OC instrument separate. Still, future work may include combinations of \change[]{concentrations}{cell abundance} maps from instruments with similar equator-crossing times and, therefore, local overpass times. OC + TROPOSIF, TROPOSIF only, and OC only outputs will not overlap by definition, so there is no need to define order or hierarchy amongst the products when combining. Along with the daily \change[]{concentrations}{cell abundance} products, we generated a Data Quality Indicator (DQI), which, for now, provides an index associated with which data stream a given pixel came from (OC +TROPOSIF, TROPOSIF only, or OC only). In the future, this product may also contain uncertainties. For this study, we also calculated monthly averages for each product and its associated DQI. This could also be done on a weekly or 8-day cadence in the future, to match other operational products.

\subsubsection{Full Prediction Pipeline}
\begin{figure*}[h!]
\noindent\centering\includegraphics[width=\textwidth]{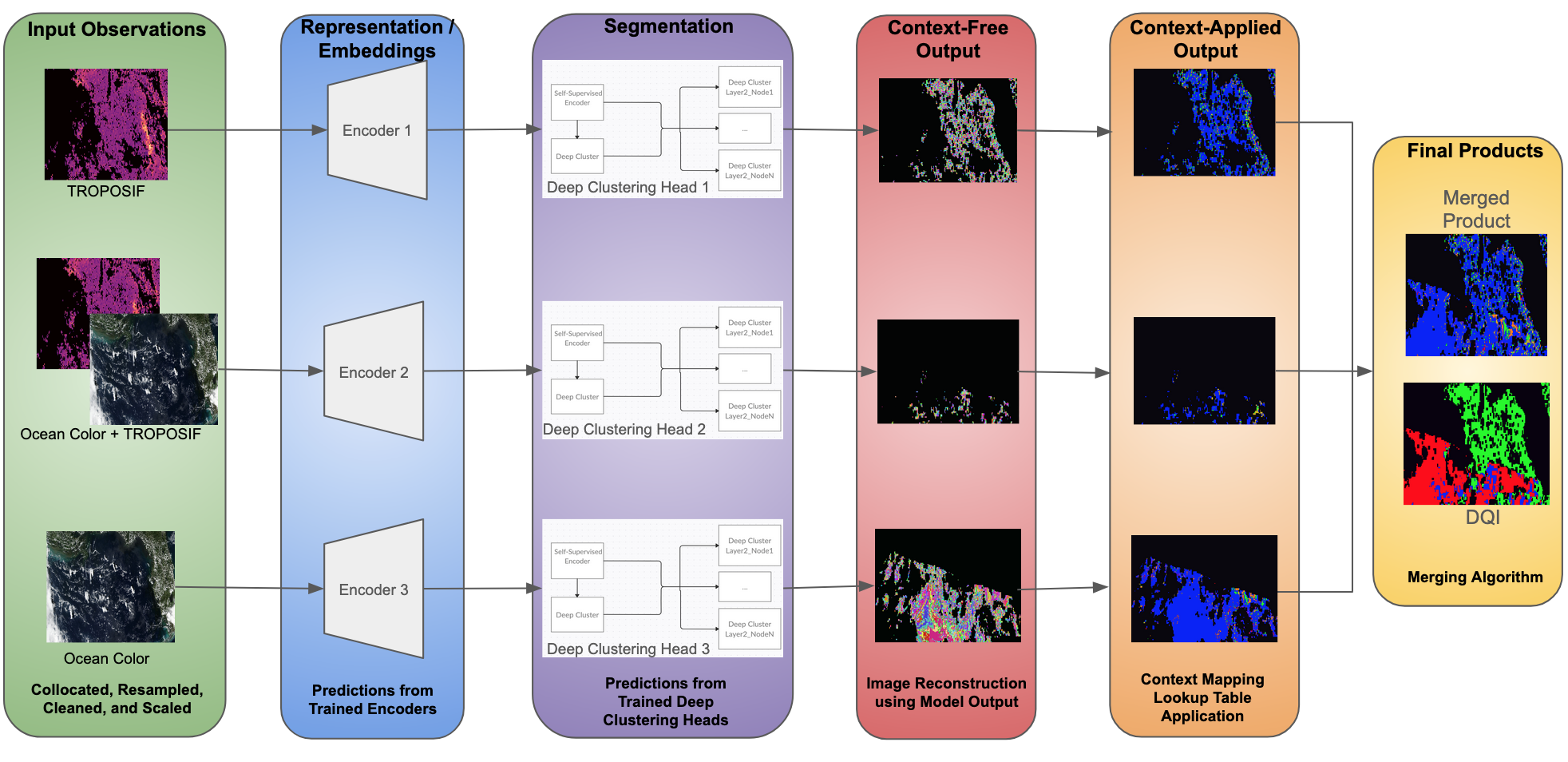}
\caption{A depiction of the full product generation pipeline for a single OC instrument + TROPOSIF data stream. Each box represents a discretized step of the process. The top of each box lists the name of the step or product generated, and the bottom lists the processes used. The three separate flows across the first four boxes depict the separate TROPOSIF, OC + TROPOSIF, and OC processing pipelines, as described above.}\label{fig_pred}
\end{figure*}

\add[]{Combining the pre-processing, trained models, context assignment, and product merging, we get the full prediction pipeline. Figure} \ref{fig_pred} \add[]{depicts the full pipeline for product generation for a single OC instrument and TROPOSIF data stream.}

\subsubsection{Validation}
\label{subsec235}
\change[]{Validation is being done in a very similar way to context assignment. Here, using the time periods held out for testing, instead of matching up the daily context-free segmentation products to the in situ sites, we are matching up the daily binned cell abundance products to the in situ sites, binned in the same way, and creating histograms to map agreement - which just become the confusion matrices provided as tables below.}{To evaluate SIT-FUSE in HAB terms, we use the same matchup strategy as for context assignment: for held-out periods, we compare the SIT-FUSE-derived abundance class at the location of each in situ station to the cell-abundance bin of the observed in situ data, and summarize agreement in confusion matrices. This directly quantifies how often the framework assigns the correct severity/species class near monitoring sites, and which misclassifications (e.g., underestimation of high-abundance events) are most common.} As shown in the results tables, there are relatively few matchups, which, as discussed before, limits the applicability of many supervised and even semi-supervised solutions \cite{SAM}. This challenge is common in the remote sensing field, \add[]{and even more pervasive in the ocean than on land.} \change[]{and}{Our} approach seeks to address it by combining self-supervised learning, the expertise of domain specialists, and the wealth of available unlabeled data \cite{Yang2019, Woodhouse2021}. By integrating these elements, we aim to develop more effective, scalable solutions to complex earth observation problems. This study provides a baseline from which we can continue to expand validation and application to larger spatiotemporal regions to better understand strengths, limitations, and related factors.

\section{Results}
\subsection{Multi-Instrument + TROPOSIF 2018 - 2019}
\label{subsec31}
\subsubsection{Gulf of Mexico}
\label{subsubsec311}

\begin{figure*}[h!]
\noindent\centering\includegraphics[width=\textwidth]{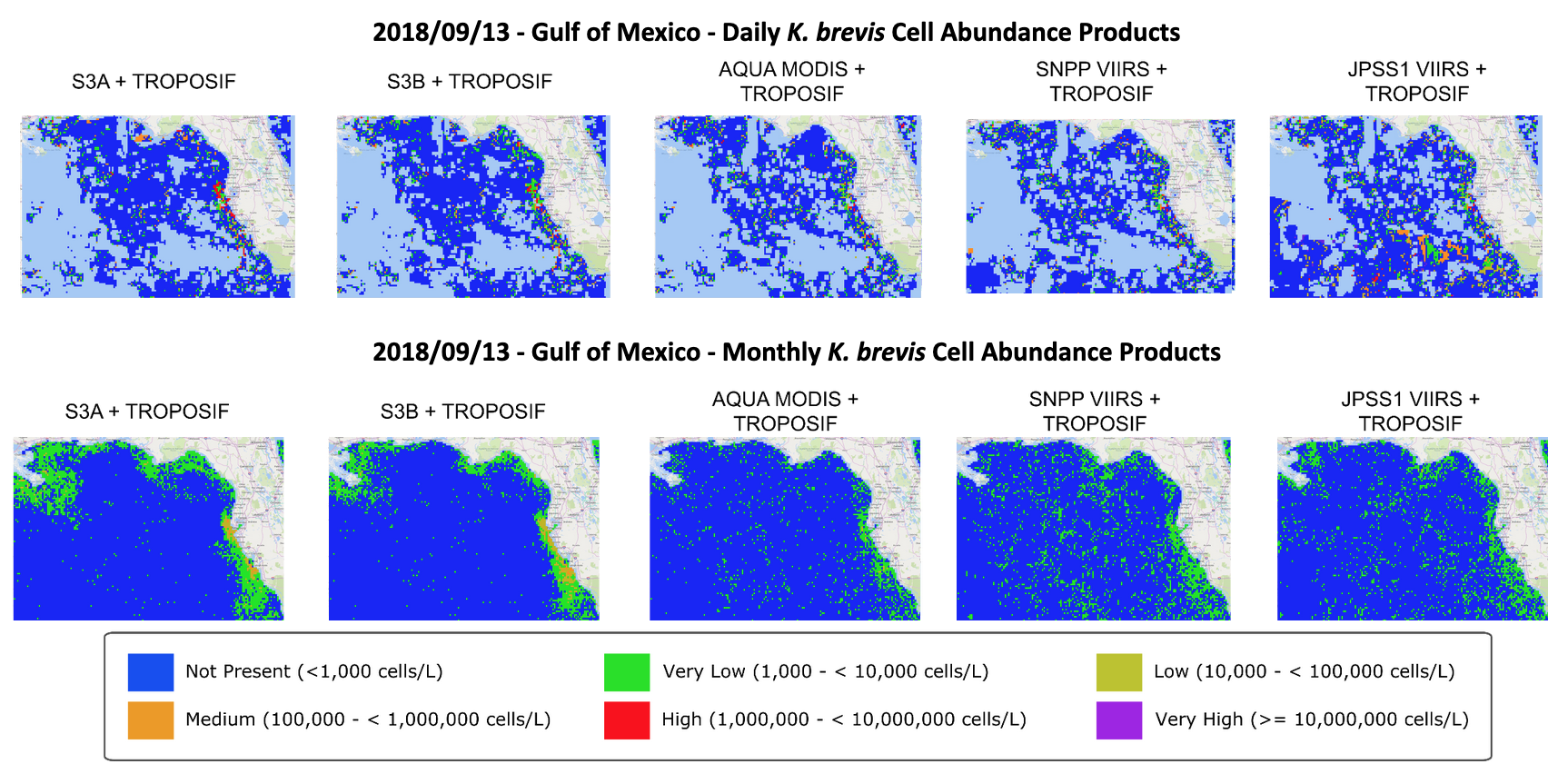}
\caption{Daily products from each of the instrument/data streams for September 13, 2018 (top) and the associated monthly products (bottom).}\label{fig6}
\end{figure*}

Table \ref{tab1} is the confusion matrix that summarizes the performance of our approach, across all input streams, for the Gulf of Mexico 2018-2019 test case, when compared to the in situ sites. The same minimum radii are set for each area of study \remove[]{(2.5km for Florida and 10km for Southern California)}. The left side of the table shows raw counts, and the right side shows the corresponding percentages. Figure \ref{fig6} depicts a single day and a single monthly average over the entire Gulf of Mexico. This time period was chosen to represent, as there was an extreme \change[]{K. brevis}{\textit{K. brevis}} bloom occurring, and it was within the period of study used in \cite{Luis2023}. \add[]{Conceptually, we ask: for each cluster produced by SIT-FUSE, which HAB cell-abundance bin does it most often coincide with at matchup locations? We then label each cluster with its corresponding abundance class (e.g., 'moderate \textit{K. brevis}'), and apply this mapping globally to all pixels with the same cluster label.}

\begin{table*}[h!]
    \centering
    \includegraphics[width=\textwidth,height=2.5cm]{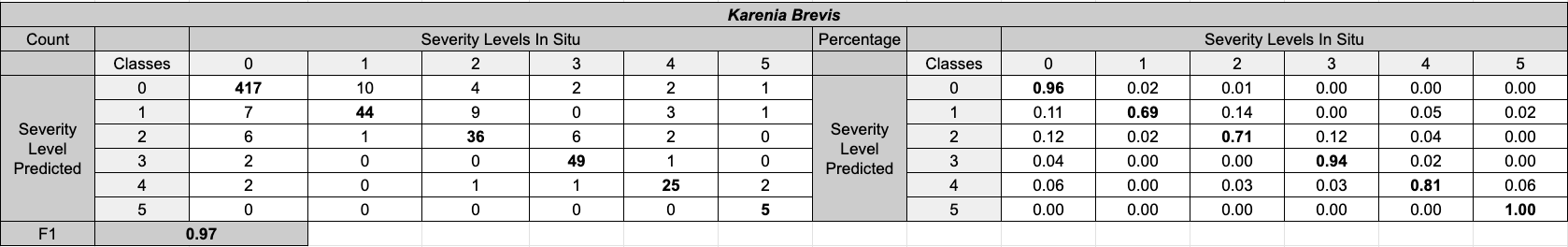}
    \caption{The comparison between binned \change[]{concentrations}{cell abundance} of \change[]{K. brevis}{\textit{K. brevis}} from in situ sites, and those predicted within the SIT-FUSE product within the scenes from the test set for the 2018-2019 Gulf of Mexico test case.  These counts encapsulate all input/output streams from the various instruments. The left table shows pixel count, and the right shows percentage. The bins are the same as those shown in Figure \ref{fig4}.}
    \label{tab1}
\end{table*}

\add[]{Overall agreement between SIT-FUSE and in-situ \textit{K. brevis} observations are strong (F1=0.97), with the highest agreement for both the lowest and highest abundance classes and slightly lower agreement for intermediate bins. The dominant off‑diagonal patterns show misclassification into adjacent abundance bins, indicating that most errors differ from the truth by only one bin. The majority of misclassification cases that do not follow this trend are false positives for the low/no presence class (1), i.e., misclassifications of pixels with a truth value of 1. These misclassifications likely reflect dynamics at the sub-pixel level that cannot be picked up at these spatial resolutions.}

\subsubsection{Southern California}
\label{subsubsec312}

\begin{figure*}[h]
\noindent\centering\includegraphics[width=\textwidth]{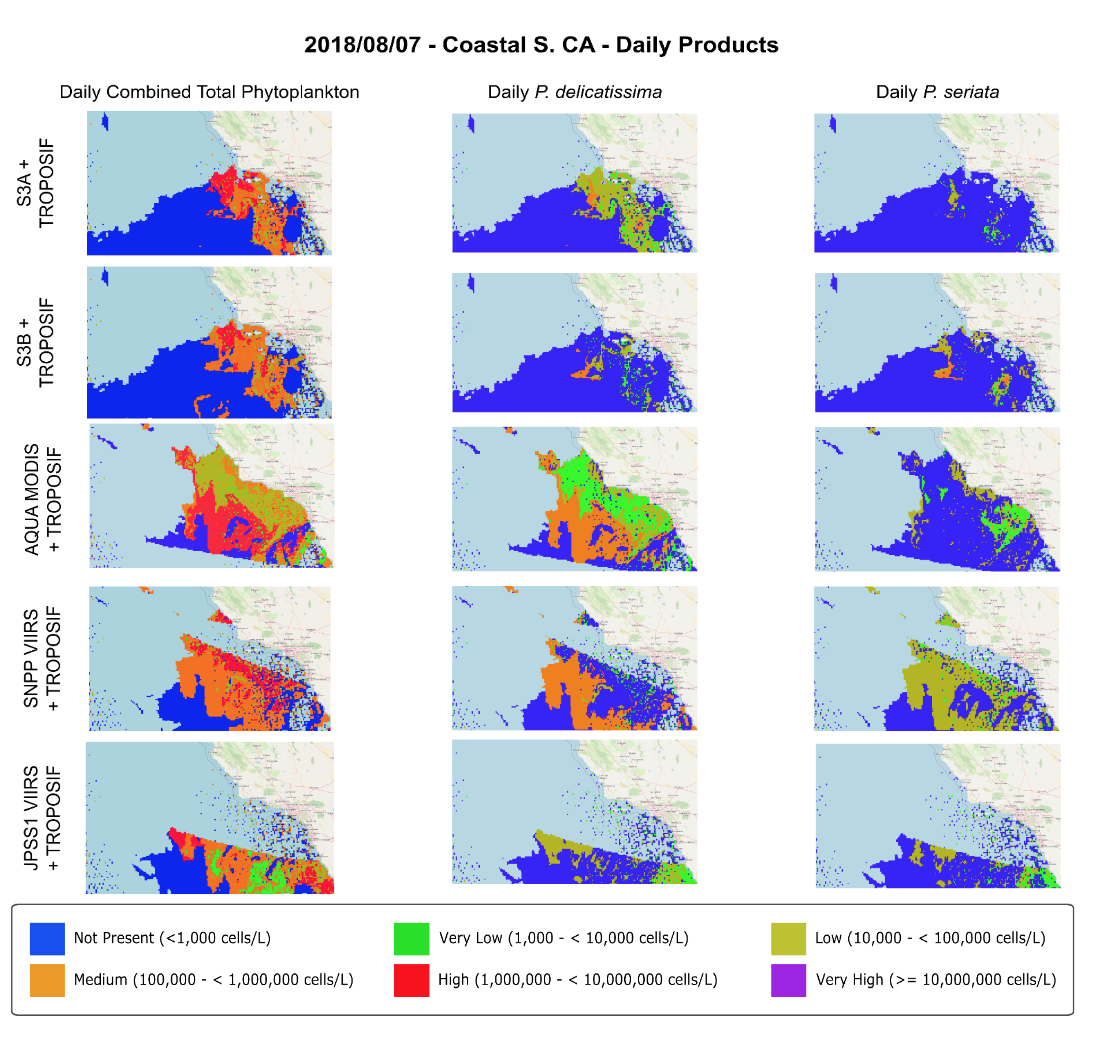}
\caption{Daily products for total phytoplankton \change[]{concentrations}{cell abundance} and each of the potential HAB forming species - \change[]{P. delicatissima}{\textit{P. delicatissima}} \change[]{, P. seriata}{\textit{P. seriata}} \remove[]{, and A. spp.} - from each of the instrument/data streams for August 7, 2018.}\label{fig7}
\end{figure*}

\begin{figure*}[h]
\noindent\centering\includegraphics[width=\textwidth]{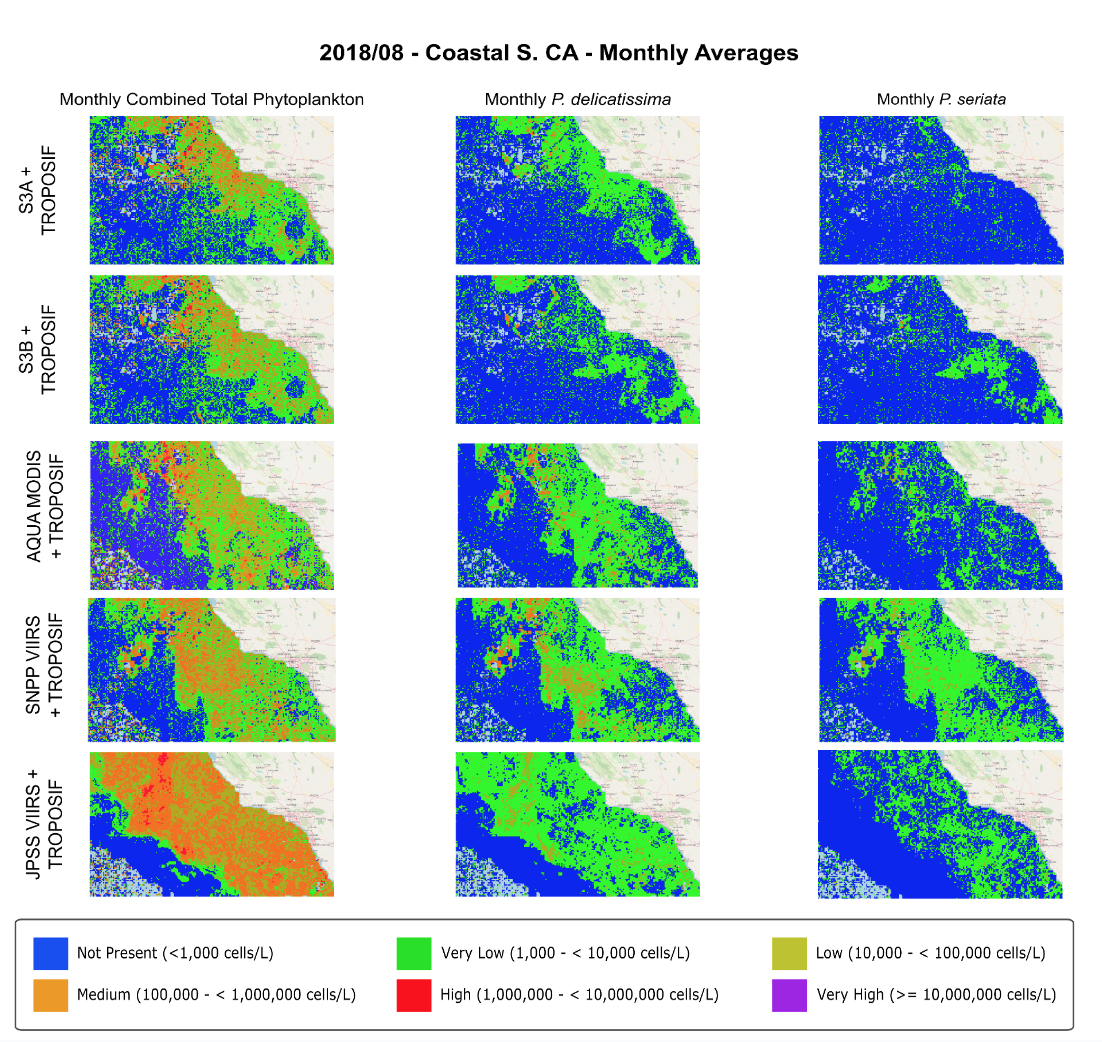}
\caption{\change[]{Daily}{Monthly} products for total phytoplankton \change[]{concentrations}{cell abundance} and each of the potential HAB forming species - \change[]{P. delicatissima}{\textit{P. delicatissima}} \change[]{, P. seriata}{\textit{P. seriata}} \remove[]{, and A. spp.} - from each of the instrument/data streams for August 7, 2018.}\label{fig8}
\end{figure*}

Table \ref{tab2} is the confusion matrix that summarizes the performance of our approach, across all input streams, for the Southern California (S. CA) 2018-2019 test case, when compared to the in situ sites. The same minimum radii are set for each area of study \remove[]{(2.5km for Florida and 10km for Southern California)}. The left side of the table shows raw counts, and the right side shows the corresponding percentages. Figure \ref{fig7} depicts a single day, and Figure \ref{fig8} depicts a single monthly average over the entire region. Tables \ref{tab2}, \ref{tab3}, and \ref{tab4} detail the results for \change[]{P. delicatissima}{\textit{P. delicatissima}} \change[]{, P. seriata}{\textit{P. seriata}} \remove[]{, and A. spp.} in the same way.

\begin{table*}[h]
    \centering
    \includegraphics[width=\textwidth,height=2.5cm]{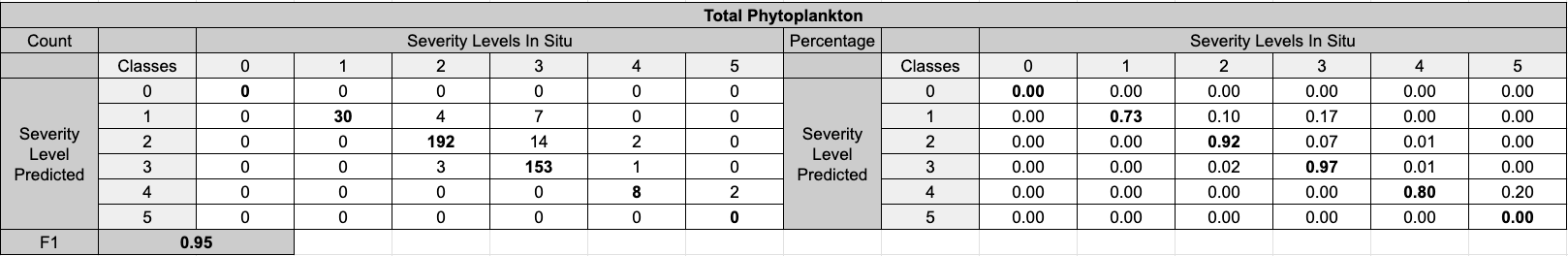}
    \caption{The comparison between binned \change[]{concentrations}{cell abundance} of total phytoplankton from in situ sites, and those predicted within the SIT-FUSE product within the scenes from the test set. For the 2018-2019 S. CA test case, these counts encapsulate all input/output streams from the various instruments. The left table shows pixel count, and the right shows percentage. The bins are the same as those shown in Figure \ref{fig4}.}
    \label{tab2}
\end{table*}

\begin{table*}[h]
    \centering
    \includegraphics[width=\textwidth,height=2.5cm]{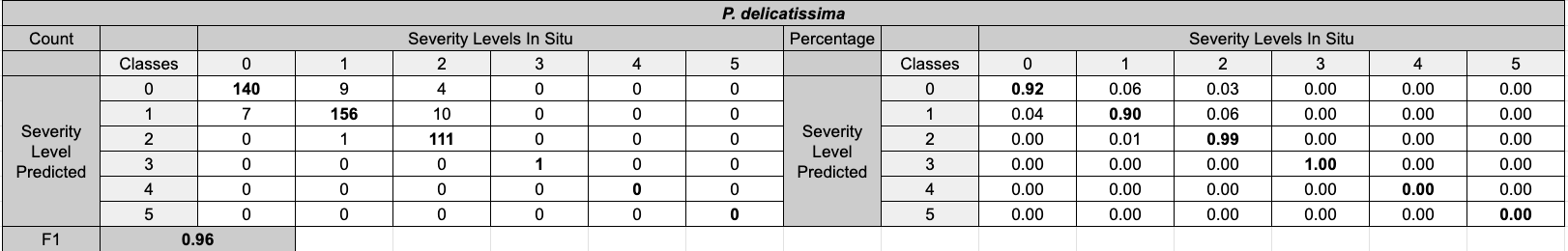}
    \caption{The comparison between binned \change[]{concentrations}{cell abundance} of  \change[]{P. delicatissima}{\textit{P. delicatissima}} from in situ sites, and those predicted within the SIT-FUSE product within the scenes from the test set for the 2018-2019 S. CA test case.  These counts encapsulate all input/output streams from the various instruments. The left table shows pixel count, and the right shows percentage. The bins are the same as those shown in Figure \ref{fig4}.}
    \label{tab3}
\end{table*}

\begin{table*}[h]
    \centering
    \includegraphics[width=\textwidth,height=2.5cm]{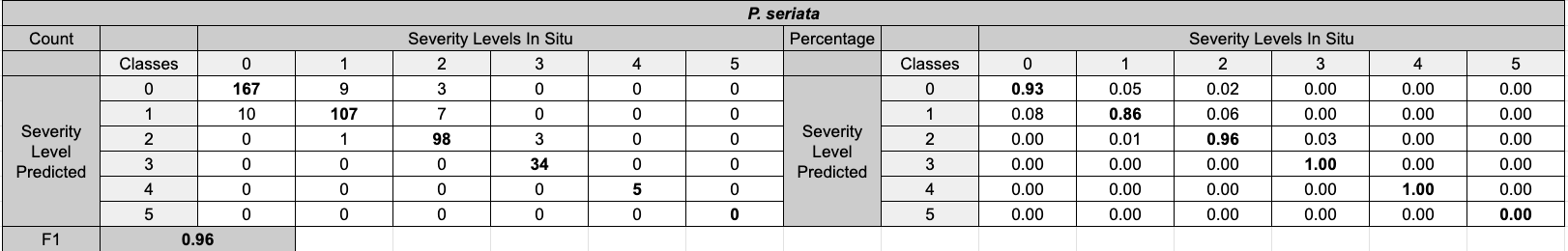}
    \caption{The comparison between binned \change[]{concentrations}{cell abundance} of \change[]{P. seriata}{\textit{P. seriata}} from in situ sites, and those predicted within the SIT-FUSE product within the scenes from the test set. For the 2018-2019 S. CA test case, these counts encapsulate all input/output streams from the various instruments. The left table shows pixel count, and the right shows percentage. The bins are the same as those shown in Figure \ref{fig4}.}
    \label{tab4}
\end{table*}

\clearpage
\add[]{Like the \textit{K. brevis} analysis, although a much smaller matchup count, overall agreement between SIT-FUSE and in-situ total phytoplankton, \textit{P. delicatissima}, and \textit{P. seriata} observations are strong (F1=0.95,96,96 respectively), with the highest agreement at the lowest and highest abundance classes. The small number of misclassifications suggests an underestimation bias, with many cases assigned to the next‑lower bin than observed.}


\subsubsection{Performance Summary}
\add[]{Across regions, SIT-FUSE achieves similarly high F1 scores for \textit{K. brevis} in the Gulf of Mexico and for total phytoplankton and \textit{P. spp.} in Southern California, but the error structure differs. In the Gulf, misclassifications are dominated by near-diagonal errors between adjacent abundance bins and occasional low-bias false positives at the lowest class, consistent with sub‑pixel variability in red tide patches. In Southern California, the few misclassifications show a more systematic underestimation of high-abundance bins, reflecting the challenge of resolving intense, nearshore blooms in optically complex, heterogeneous coastal waters.}
\subsection{A first look at PACE-based Retrievals 2024 - 2025}
\label{subsec32}
\subsubsection{Gulf of Mexico}
\label{subsubsec321}

\begin{figure*}[h]
\noindent\centering\includegraphics[width=\textwidth]{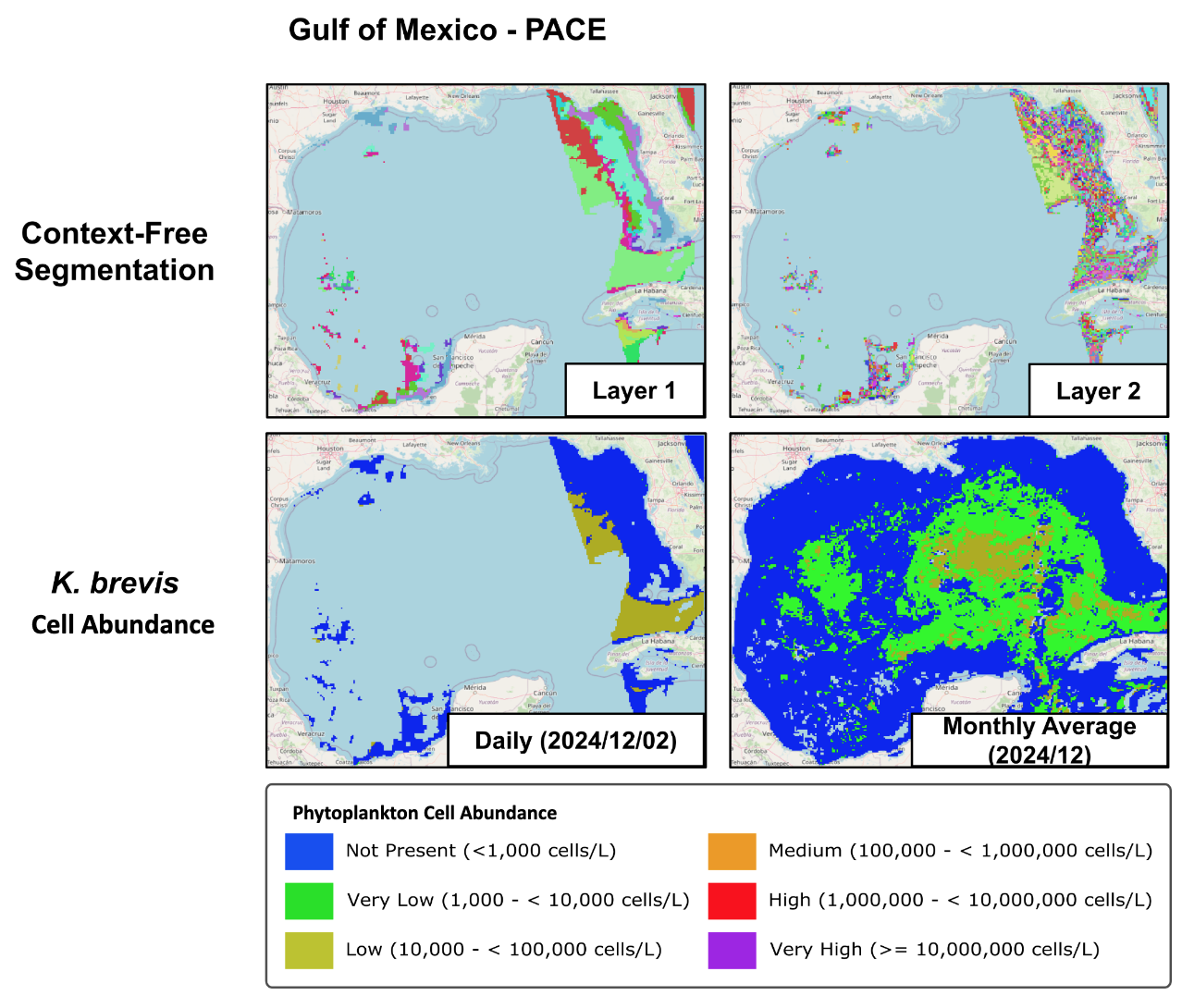}
\caption{Daily context free segmentation products (top) and \change[]{K. brevis}{\textit{K. brevis}} \change[]{concentrations}{cell abundance} (center) generated from PACE data for December 2, 2024, and the associated monthly products (bottom). The Florida Department of Health in Collier County cautioned the public about a red tide near Clam Pass and Barefoot Beach in response to a water sample taken on December 5, 2024; however, the spatial extent of the event and its nearshore proximity, along with the lack of matchups, rendered it undetectable by PACE.  \change[]{concentrations}{cell abundance} legend can be found in Figure \ref{fig4}.}\label{fig9}
\end{figure*}

Table \ref{tab6} is the confusion matrix that summarizes the performance of our approach using PACE OCI reflectances, for the Gulf of Mexico 2024-2025 test case, when compared to the in situ sites. The exact minimum radii are set for each area of study \remove[]{(2.5km for Florida and 10km for Southern California)}. The left side of the table shows raw counts, and the right side shows the corresponding percentages. Figure \ref{fig9} depicts the generated products for a single day and a single monthly average over the entire Gulf of Mexico. While the counts in the PACE Gulf of Mexico case are too low to conduct a proper quantitative evaluation of performance, we feel it is worthwhile to demonstrate progress in this direction. We will add more data to this evaluation as PACE gathers more data over this region. Table 6 presents comparisons across the matchups we had access to in the test set. 

\begin{table*}[h]
    \centering
    \includegraphics[width=\textwidth,height=2.5cm]{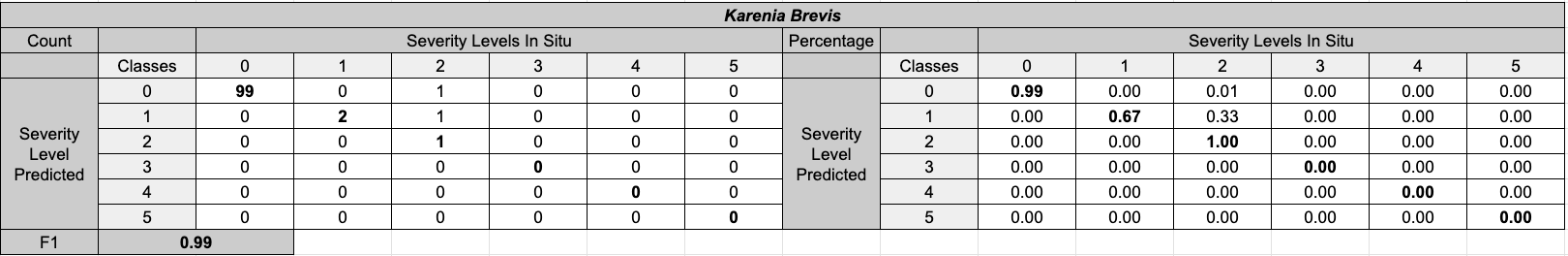}
    \caption{The comparison between binned \change[]{concentrations}{cell abundance} of \change[]{K. brevis}{\textit{K. brevis}} from in situ sites, and those predicted within the SIT-FUSE product within the scenes from the test set. For the 2024-2025 Gulf of Mexico PACE test case, the left table shows pixel count, and the right shows percentage. The bins are the same as those shown in Figure \ref{fig4}. Counts are far too low to provide a good evaluation; more data will be added to this evaluation as PACE continues to collect data.}
    \label{tab6}
\end{table*}

\add[]{Within the small observation set, agreement is strong (F1=0.99), a promising first signal. Minimal variance and a small sample size do not allow for further insight at this time.}
\subsubsection{Southern California}
\label{subsubsec322}

\begin{figure*}[p]
\begin{minipage}{\textwidth}
    \centering
    \includegraphics[width=\textwidth,height=2.5cm]{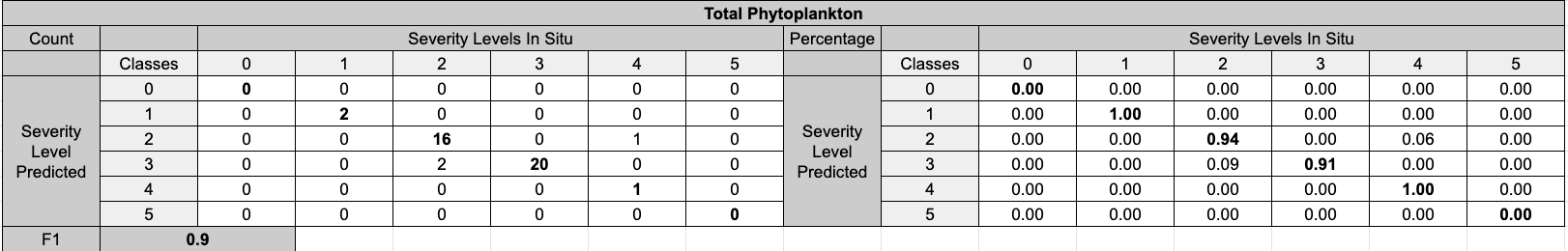}
    \captionof{table}{The comparison between binned \change[]{concentrations}{cell abundance} of total phytoplankton \change[]{concentrations}{cell abundance} from in situ sites, and those predicted within the SIT-FUSE PACE product within the scenes from the test set for the 2024-2025 S. CA test case.}
    \label{tab7}
 \end{minipage}
    \hfill
     \vspace{1cm}
\begin{minipage}{\textwidth}
    \centering
    \includegraphics[width=\textwidth,height=2.5cm]{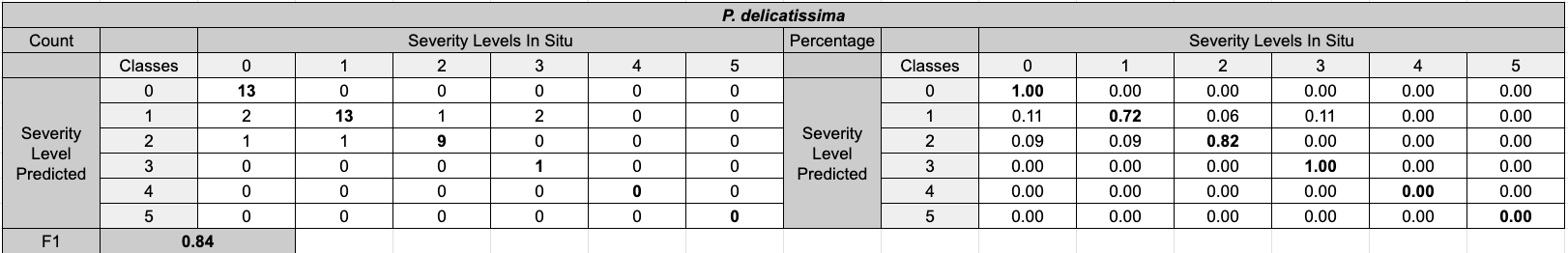}
    \captionof{table}{The comparison between binned \change[]{concentrations}{cell abundance} of  \change[]{P. delicatissima}{\textit{P. delicatissima}} from in situ sites, and those predicted within the SIT-FUSE PACE-based total phytoplankton product within the scenes from the test set for the 2024-2025 S. CA test case.}
    \label{tab8}
 \end{minipage}
    \hfill
     \vspace{1cm}
  \begin{minipage}{\textwidth}
    \centering
    \includegraphics[width=\textwidth,height=2.5cm]{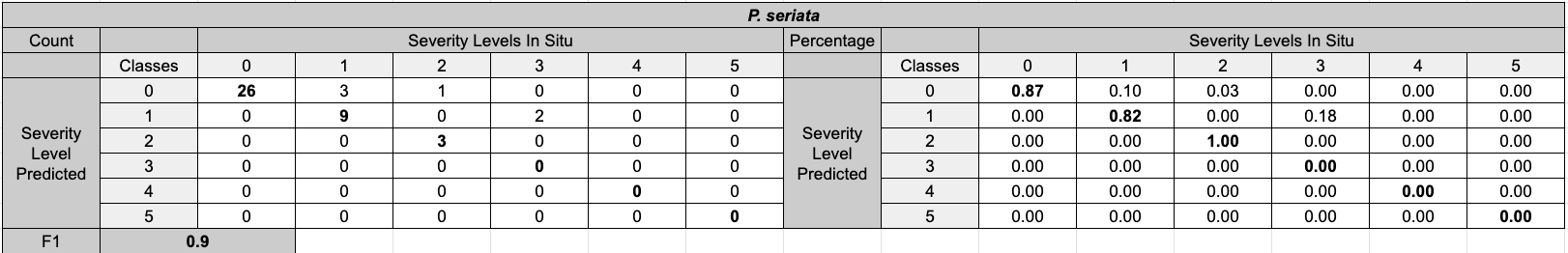}
    \captionof{table}{The comparison between binned \change[]{concentrations}{cell abundance} of \change[]{P. seriata}{\textit{P. seriata}} from in situ sites, and those predicted within the SIT-FUSE PACE-based product within the scenes from the test set for the 2024-2025 S. CA test case. }
    \label{tab9}
 \end{minipage}
    \hfill
     \vspace{1cm}
  \begin{minipage}{\textwidth}
\noindent\centering\includegraphics[width=\textwidth]{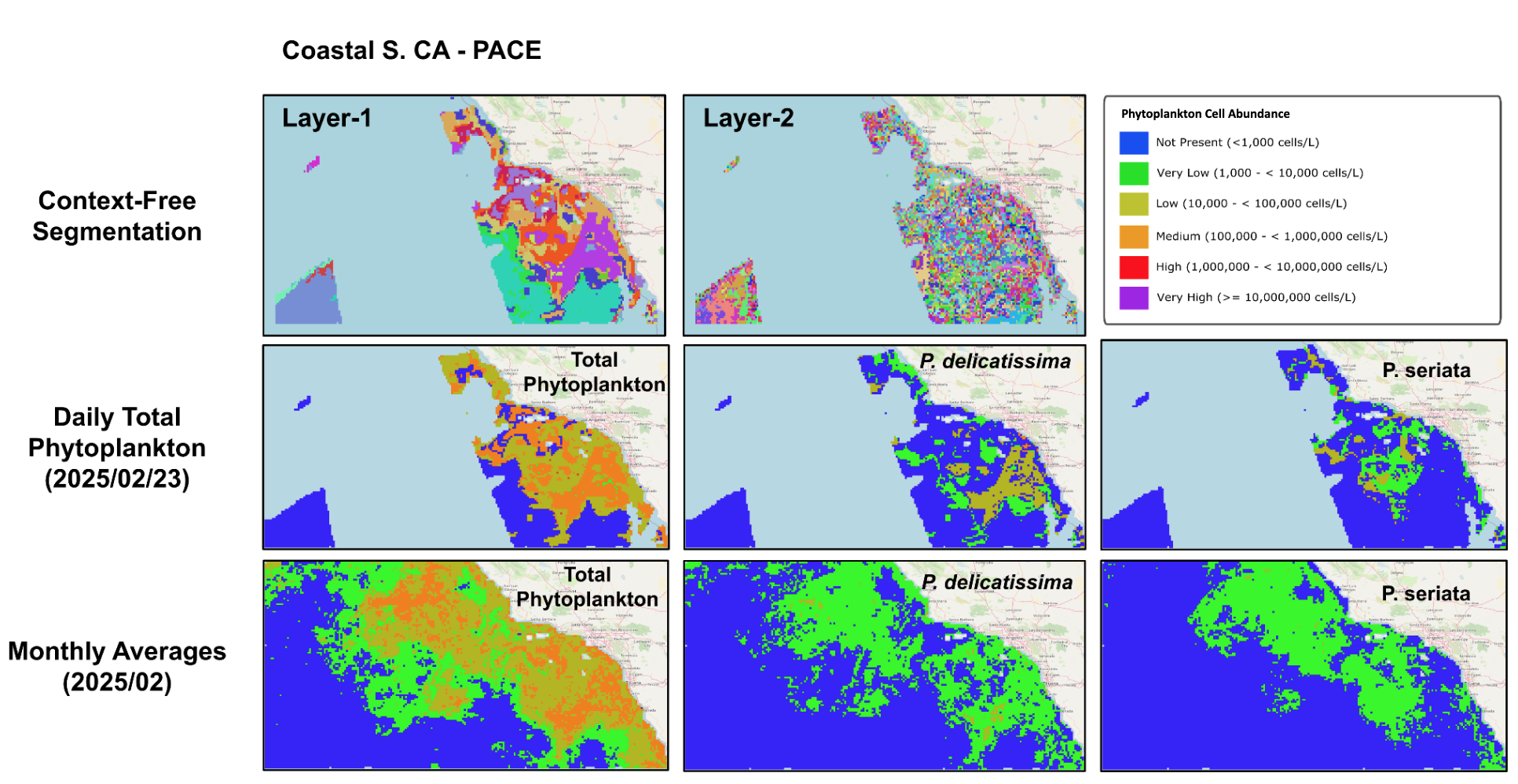}
\caption{Daily context free segmentation products (top), total phytoplankton \change[]{concentrations}{cell abundance} (row 2), and speciated HAB \change[]{concentrations}{cell abundance} (row 3), generated from PACE data for December 2, 2024, and the associated monthly total \change[]{concentrations}{cell abundance} product (bottom). \change[]{concentrations}{cell abundance} legend can be found in Figure \ref{fig4}.}\label{fig10}
 \end{minipage}
\end{figure*}

Table \ref{tab7} presents the confusion matrix summarizing the performance of our approach using PACE for the S. CA 2024-2025 test case, compared to the in situ sites. The same minimum radii are set for each area of study \remove[]{(2.5km for Florida and 10km for Southern California)}. The left side of the table shows raw counts, and the right side shows the corresponding percentages. Figure \ref{fig10} depicts a single day and a single monthly average over the entire region. Tables \ref{tab8} and \ref{tab9} detail the results for \change[]{P. delicatissima}{\textit{P. delicatissima}} and \change[]{P. seriata}{\textit{P. seriata}} in the same way. \remove[]{There was too little variation in \textit{A. spp.} concentration over this time period to generate products and do similar evaluations to the section}
\clearpage
\add[]{Here again we see promising signals in this first look at performance with PACE cell abundance classification for total phytoplankton, \textit{P. delicatissima}, and \textit{P. seriata} and (F1=0.9, 0.84, and 0.9 respectively). The small number of misclassifications appears to be forming patterns similar to those in the multi-sensor analysis, but again, sampling a larger window of available PACE observations will help us make stronger determinations.}

\subsubsection{Performance Summary}
\add[]{PACE‑based products show promising first‑look skill, with high F1 scores for all cases, despite very limited matchups, but the small sample sizes preclude detailed assessment of sensor‑specific biases. The emerging pattern of near-diagonal misclassification and occasional underestimation of high‑abundance events mirrors the multi‑sensor results, suggesting that this methodology also functions well in the hyperspectral instrument domain. As PACE accumulates more observations, these preliminary evaluations can be extended to quantify whether its finer spectral resolution reduces underestimation in high‑severity coastal blooms.}

\subsection{Further Analysis}
\label{subsec33}
\subsubsection{Performance Distribution}
\begin{figure*}[h!]
\noindent\centering\includegraphics[width=\textwidth]{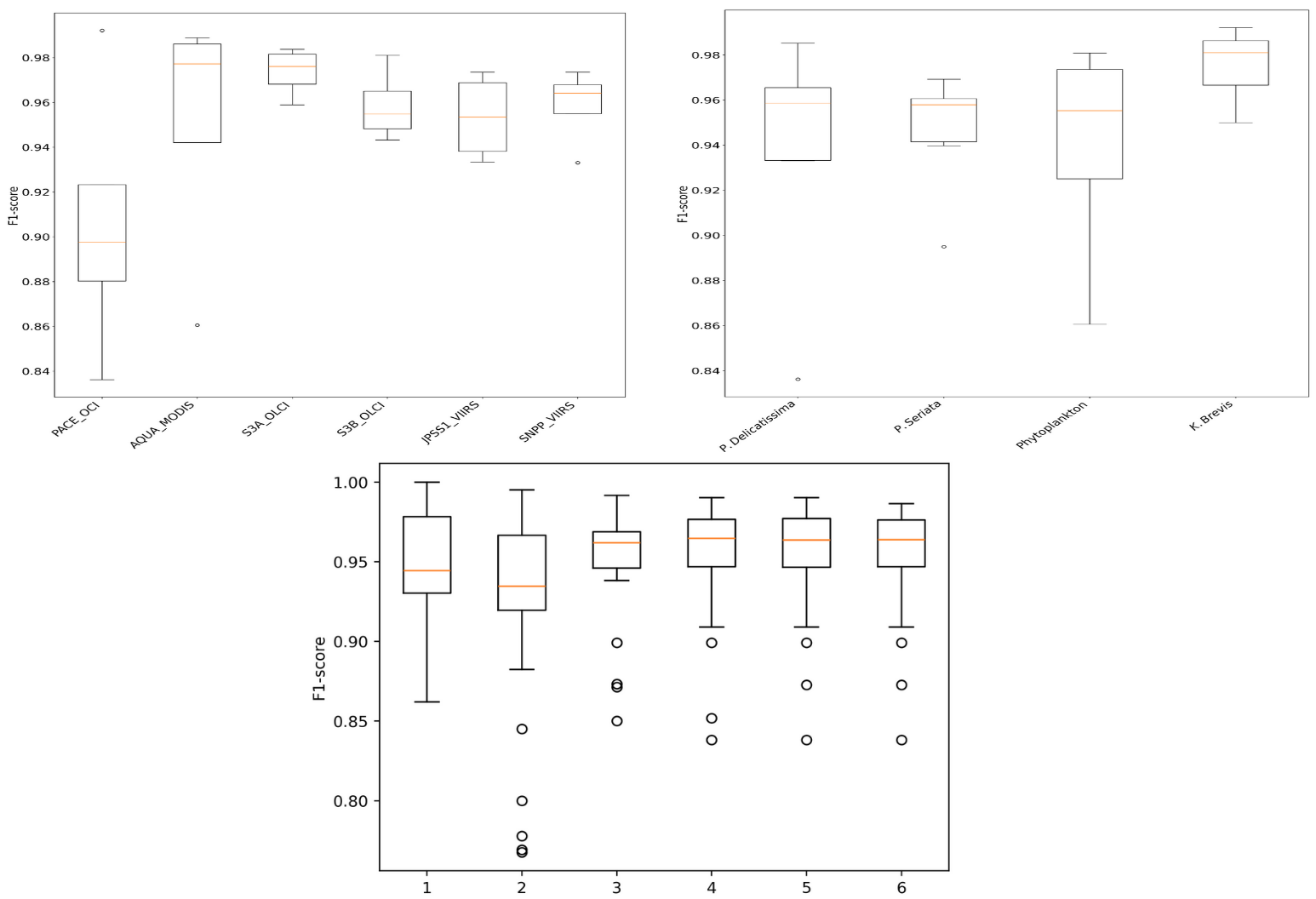}
\caption{Boxplots of F1 distributions for all experiments, split by instrument (top left), phytoplankton product (top right), and cell abundance bin (bottom). All median values and interquartile ranges are quite high, within the range of $[0.85 ,0.98]$, with the largest outliers falling slightly below 0.8.}\label{fig11}
\end{figure*}
\add[]{While it is clear that the true positives dominate the confusion matrices, and the associated F1-scores are high, we take a closer look at different breakdowns of these values. Performance was measured per-instrument, per-cell-abundance-level, and per-species/product, and Figure} \ref{fig11} \add[]{depicts those results. We can see that the distributions of F1 scores are high and relatively similar across all instruments, with the lowest scores coming from PACE, the instrument with the fewest samples in this study. With respect to species/product, the distributions are again high and relatively similar, with the total phytoplankton product showing a much wider spread than the others. Lastly, with respect to cell abundance levels, it looks like the widest spread/variance occurs in the lower ranges. This also shows up clearly in the confusion matrices and is likely due to higher sample sizes and, in some cases, the difficulty of discerning between no and low cell abundance. Overall, the performance in this study, both in general and split by instrument, product, and cell abundance level, is high, indicating effective representation of phytoplankton distributions near in-situ sites.}

\subsubsection{Comparison to Operational Models}
\begin{figure*}[h!]
\noindent\centering\includegraphics[width=\textwidth]{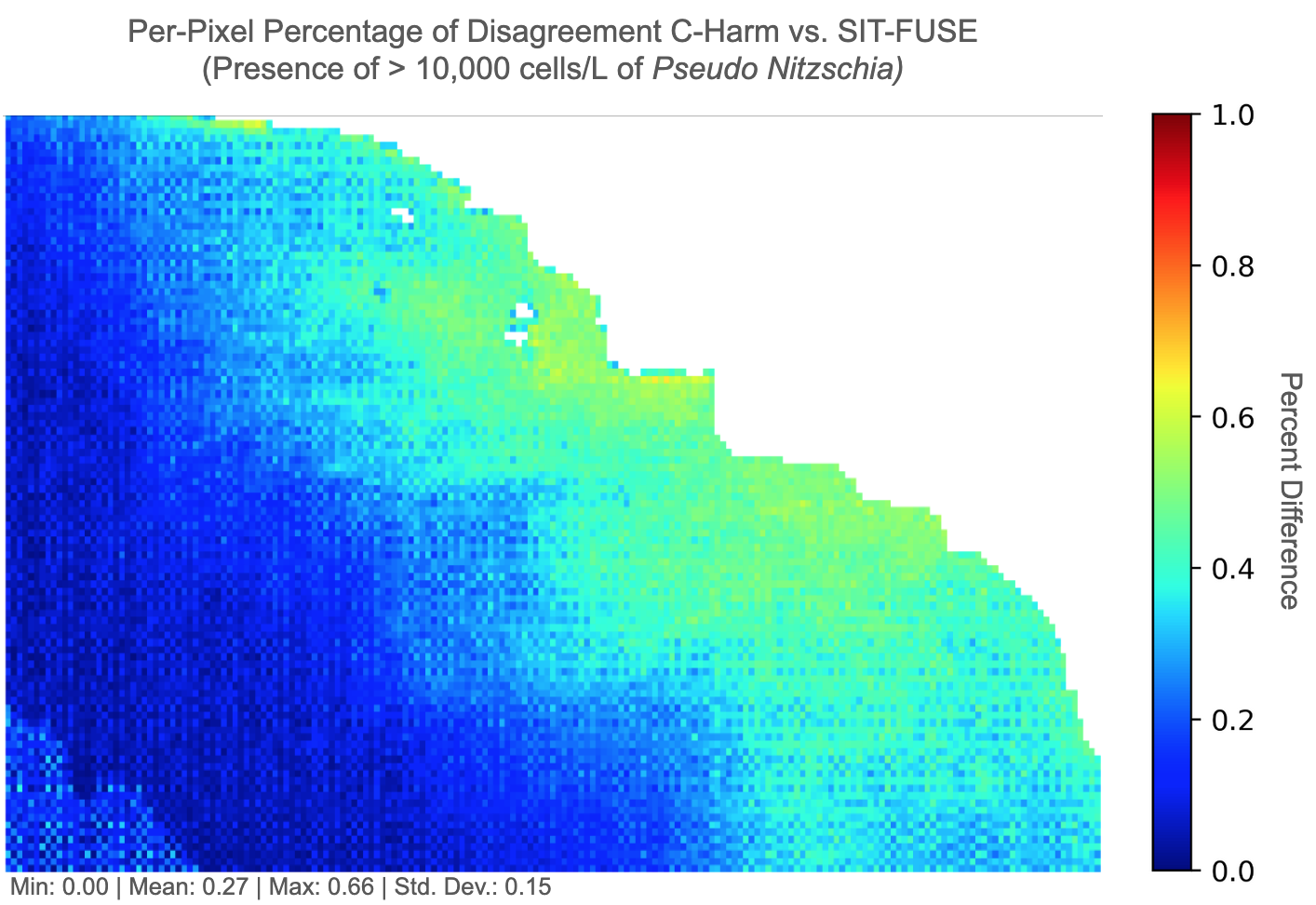}
\caption{Plot displaying the per-pixel \% difference between binarized SIT-FUSE \textit{P. spp.} products and binarized C-HARM nowcasts for probability of \textit{P. spp.} $>10000$ cells/L. Summary statistics for the differences are also available at the bottom left.}\label{fig11a}
\end{figure*}

\add[]{To get an idea of performance in a wider spatial context in Southern California, and do some initial comparisons in areas where we have no in situ observations, we compare against a model that is currently in use operationally. The California-Harmful Algae Risk Mapping (C-HARM) is designed to predict HABs caused by the diatom \textit{P. spp}. and its neurotoxin, domoic acid (DA), along the U.S. West Coast. Developed through collaborations between NOAA, NASA, and regional ocean observing systems (SCCOOS, CeNCOOS), C-HARM integrates physical circulation models (e.g., Regional Ocean Model System/ROMS) to simulate ocean temperature, salinity, and currents, satellite remote sensing (MODIS-Aqua) for ocean color, chlorophyll, and optical parameters, and statistical ecological models to estimate bloom and toxin probabilities} \cite{Anderson2016} \add[]{.  C-HARM generates daily nowcasts and 3-day forecasts for \textit{P. spp} Bloom Probability - likelihood of exceeding 10,000 cells/L, a threshold linked to toxin production, particulate domoic acid (DA) risk - probability of DA cell abundance $\geq$ 500 ng/L in phytoplankton, and cellular toxicity - probability of DA $\geq$ 10 pg/cell in \textit{P. spp}, indicating highly toxic cells. Like the output of this project, the model’s skill has been validated against nearshore monitoring data from the California HAB Monitoring and Alert Program (HABMAP), with high agreement closer to shore and some discrepancies farther offshore, highlighting the need for ongoing offshore sampling. Recent iterations (e.g., C-HARM v3) incorporate the West Coast Operational Forecast System (WCOFS) for improved accuracy} \cite{Moreno2022}. \add[]{As a first step in intercomparisons between C-HARM nowcast products and the products produced via SIT-FUSE, we take both products and threshold them to make binary products. Because this initial version of the SIT-FUSE products does not have a probabilistic form, we need to discretize the C-HARM output to compare it properly. To do this, we set the threshold to 0.75. Anything $\geq 0.75$ gets labeled as a positive label, indicating the presence of \textit{P.spp.} $> 10000$ cells/L. For the SIT-FUSE products, to binarize, we set all values $\geq 2$ to 1, since 10000 cells/L is the lower bound for this class's bin. The C-HARM product was also resampled from 4km to 7km to match the SIT-FUSE products' resolution. Given that the data are binary at this point, nearest-neighbor resampling was used. All samples were taken for cases where both C-HARM and SIT-FUSE have an output value, and an F1-score was computed to measure agreement between the two. This was repeated on a per-instrument basis. Table} \ref{tab10} \add[]{depicts these results. To visualize the spatial variability of agreement, for each pixel we computed the $\%$ disagreement relative to the number of times that pixel contained valid data from SIT-FUSE and C-HARM. Figure} \ref{fig11a} \add[]{depicts these results. We can see that the general agreement is high, given that the F1 scores are between 0.72 and 0.77, and the mean $\%$ disagreement per-pixel is $27\%$, with a standard deviation of $0.15\%$. Variability increases as we move closer to shore, which makes sense given the higher variability of both the underlying system and inputs to both C-HARM and SIT-FUSE as they approach the coastline. 

While this study primarily aims to demonstrate capabilities, continued work to gather additional samples will help improve understanding of this approach's strengths and limitations at the large-scale level. Future work will take a deeper look at characterizing agreements and differences between operational products and models like C-HARM and SIT-FUSE, helping further define where a system like this best complements what is currently available operationally.}

\remove[]{As a qualitative comparison, we have overlaid our example cases of MODIS + TROPOSIF and PACE over the Nowcast of P. spp bloom probability in Figures, and it appears that there is significant agreement.

Also, Chl-a is widely used as a proxy for phytoplankton concentrations and biomass in aquatic ecosystems. This pigment is present in all photosynthetic phytoplankton and is essential for capturing light energy during photosynthesis. Because Chl-a is a common and quantifiable component of phytoplankton cells, its concentrations serve as a convenient indicator of the amount of phytoplankton present in a water sample. Given this, we also present a qualitative comparison between our concentration product for the same cases and the cumulative Chl-a products from instruments with an overpass time close to 1:30 pm local time in the figures. Again, there is significant agreement between areas with noted concentrations of Chl-a and areas our product identifies as containing high concentrations of total phytoplankton P. spp.} 

\begin{table*}[h!]
    \centering    \includegraphics[width=\textwidth,height=1.5cm]{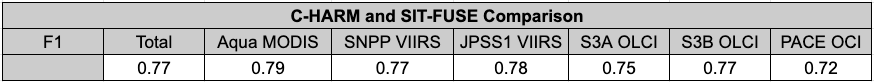}
    \caption{A table displaying the total F1-score and per-instrument F1 scores for the comparison between binarized SIT-FUSE \textit{P. spp.} products and binarized C-HARM nowcasts for probability of \textit{P. spp.} $> 10000$ cells/L.}
    \label{tab10}
\end{table*}

\section{Current and Future Work}
\label{sec4}
\subsection{Per-pixel certainties}
\label{subsec41}
\begin{figure*}[h!]
\noindent\centering\includegraphics[width=\textwidth]{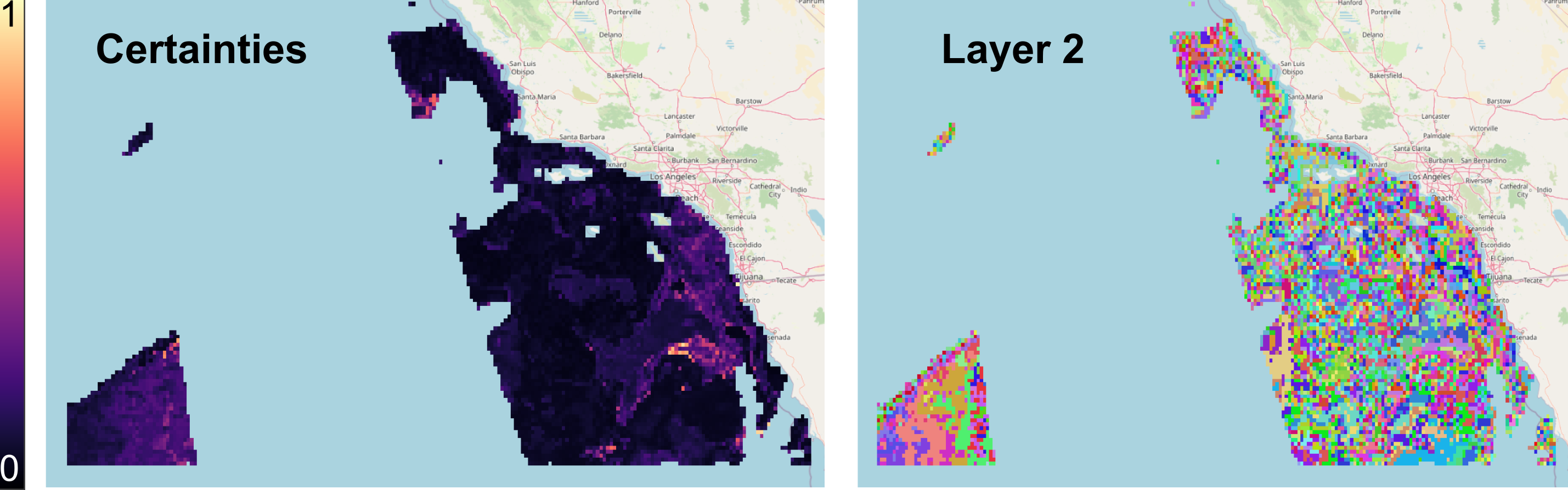}
\caption{Preliminary per-pixel certainty (left) from the lowest layer (layer 2) of the context-free segmentation generation, alongside the associated segmentation product (right).}\label{fig13}
\end{figure*}

To quantify per-scene uncertainties, we can pass forward the model's prediction scores, as shown in Figure \ref{fig13}, which can serve as a proxy for the network’s prediction uncertainty if the network is calibrated correctly \cite{Guo2017}. Various calibration techniques are currently being evaluated, so future versions of datasets produced from SIT-FUSE will also provide per-pixel information for all scenes. Given this per-pixel information, downstream applications and users can leverage uncertainties along with multi-class masks. Also, given the hierarchical nature of the context-free segmentation, we are looking into ways to propagate probabilities from each layer along the way and to evaluate downstream utility relative to just providing information from the penultimate layer. \add[]{At present, these prediction scores are uncalibrated and are therefore best interpreted as relative, not absolute, measures of confidence; ongoing calibration work is needed before they can support probabilistic decision thresholds in an operational setting.}

\subsection{Extensions to other instruments and water bodies}
\label{subsec42}

\begin{figure}[h!]
\noindent\centering\includegraphics[width=\textwidth]{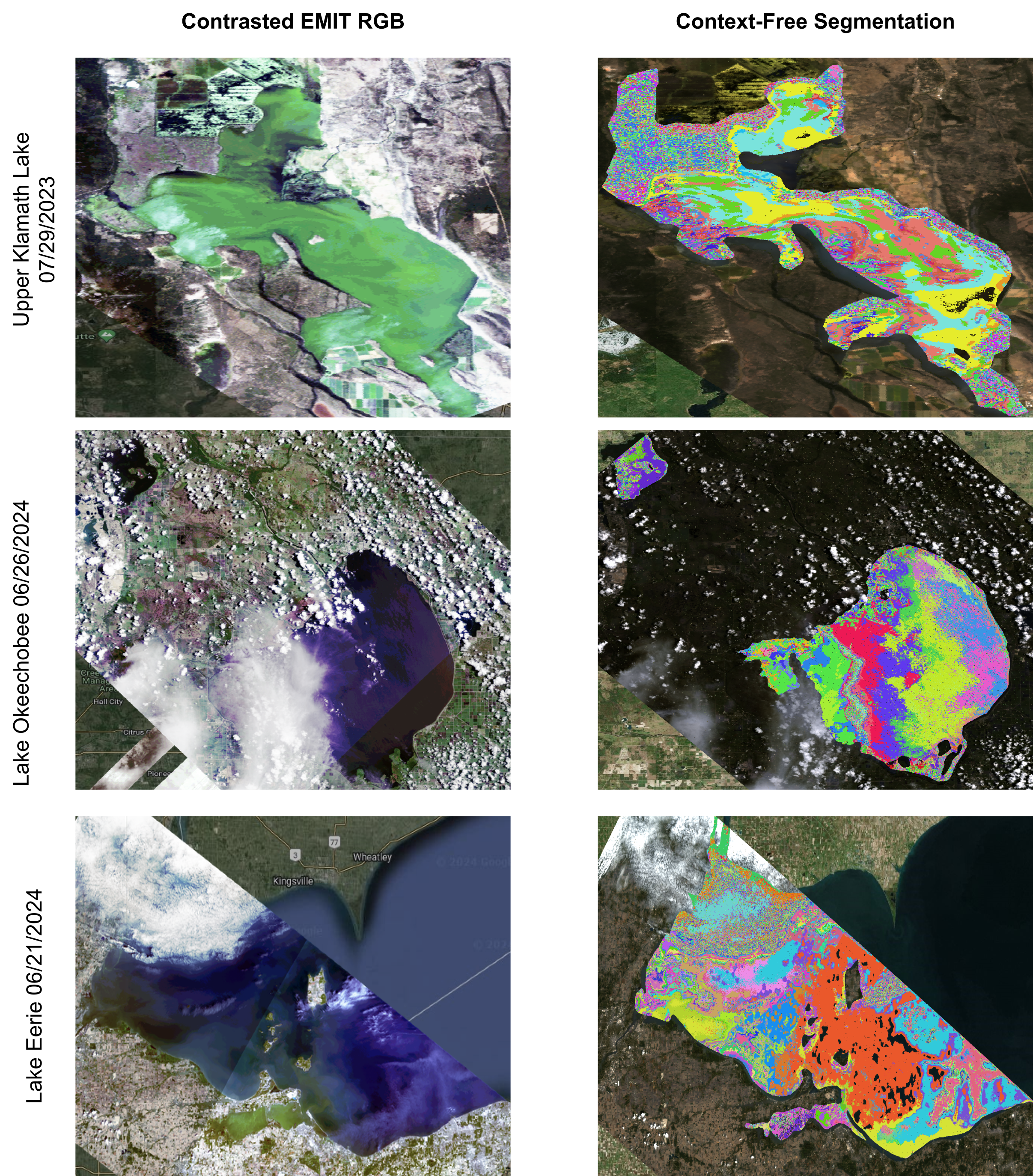}
\caption{Preliminary SIT-FUSE-based context-free segmentation products from EMIT scenes containing inland water body HABs.}\label{fig14}
\end{figure}

Increases in spatial and temporal coverage are currently underway, including expansions for the entire U.S. coastline, adding the 2020-2024 time period to the analysis, and expanding the analysis to inland water bodies. Along with these goals, as we aim to create an ad hoc sensor web from pre-existing instruments to generate datasets that enable tiered and hierarchical analysis of HAB systems, other instruments with greater variance in spatial, spectral, and temporal resolution are also of interest. \add[]{For nearshore and inland water bodies that are too fine-scaled to be caught by PACE, we aim to be able to do this approach with higher-spatial resolution imaging spectrometers, like EMIT. We also aim to obtain higher-temporal-resolution evolutions from geostationary instrumentation such as GOES.} With the first of these goals in mind, initial work has been done to use EMIT data to generate context-free segmentations over various algal bloom events in inland water bodies in the U.S. in  Figure \ref{fig14}. \add[]{While next steps include detailed quantitative analysis over the spatiotemporal regions of interest and an updated processing pipeline to account for tiered, but interconnected products, we feel this is an important first step in that direction.} 

While EMIT advances capabilities in spatial and spectral resolution, geostationary instrumentation, such as the ABIs onboard the GOES satellite platforms, can provide crucial information about diurnal cycles at fine temporal resolutions. With recent demonstrations of chl-a \change[]{concentrations}{cell abundance} from GOES ABIs \cite{Zheng2023}, and success with applying SIT-FUSE to GOES data to detect and track fires and smoke plumes,  it is also of interest to this team to test the efficacy of adding these instruments to this body of work as well.

\clearpage
\section{Conclusion}
The SIT-FUSE framework demonstrates robust capability to identify, classify, and speciate phytoplankton blooms by fusing multispectral and hyperspectral reflectance data with red SIF measurements from Sentinel-5P/TROPOMI, when available. Initial validation shows potential to enhance the temporal resolution and specificity of phytoplankton products, including HAB severity mapping, through dynamic integration of diverse sensor data. We achieve this by developing an ad hoc sensor web of instruments for phytoplankton \change[]{concentrations}{cell abundance} mapping and speciation, and by developing brand-new products for the instruments used in this experiment set. These findings demonstrate considerable promise for directly applying this technique within the domain, not only for immediate segmentation and instance tracking tasks but also for broader product development. The approach enables dynamic tracking of instances across scenes that share similar input data, enhancing flexibility and robustness. Furthermore, by integrating style transfer methods, it is likely possible to extend instance-tracking capabilities to multi-sensor scenes originating from disparate datasets, supporting cross-domain generalization and improved adaptability for complex earth observation challenges.

This methodology enables the integration of both single- and multi-instrument datasets to produce more detailed and comprehensive maps of HAB severity, improving spatial, spectral, and temporal resolution across the study area. By using encoder-derived, embedding-based representations, the framework provides a consistent, unified view of the input data and the resulting clusters. The spatial distribution of these cluster labels can then be leveraged to effectively track HAB events, even across scenes captured by different sensor systems and spanning diverse spatiotemporal domains.

For feature interpretability and selection, techniques such as embedding analysis, SHAP (Shapley Additive Explanations), and other model explainability tools can be instrumental in assessing feature importance and refining representational accuracy. Applying these methods enables a more targeted focus on spectral bands most effective for detecting key phytoplankton characteristics. Given SIT-FUSE’s strong performance, particularly in scenarios lacking pre-existing operational methodologies for HAB severity or speciation, this framework can be incorporated into new or existing instrumentation pipelines. This integration allows SIT-FUSE to replace or complement traditional instrument-specific retrieval algorithms, whose development often demands significant resources. Additionally, SIT-FUSE's segmentation capabilities deliver further efficiencies: by isolating relevant objects, the data volume required for downstream phytoplankton retrievals is significantly reduced. As a result, only the pixels corresponding to detected features require detailed processing, streamlining, and optimizing the analysis pipeline.

We have designed the SIT-FUSE framework to support a diverse range of encoder architectures flexibly, enabling systematic analysis of how different model types, complexities, and training paradigms perform in earth observation tasks. As the field continues to produce new architectures, including large Earth Observation Foundation Models (EOFMs), it becomes increasingly important to assess the representational capabilities of various encoder designs. SIT-FUSE facilitates comparative studies of models ranging from deep belief networks (DBNs) to EOFMs, enabling evaluation across a variety of conditions, problem domains, and input datasets. This adaptability ensures the framework remains relevant as foundational technologies in earth observation evolve \cite{Marsocci2024}. While assessing downstream task performance provides valuable insights, it alone does not fully characterize model capability. Recent advances, especially in large language models (LLMs), have introduced more comprehensive strategies for evaluating representation quality. Many of these robust evaluation methodologies can be adapted to computer vision and, in the context of Earth Observation, offer richer insights into a model’s ability to generalize and encode meaningful features \cite{Duderstadt2023}. Leveraging SIT-FUSE's adaptable infrastructure, we are actively developing foundational approaches to address these open challenges.

Lastly, we are actively applying SIT-FUSE to advance scientific analysis and understanding of phytoplankton \remove[]{dynamics} and harmful algal blooms (HABs). The hierarchical, context-free segmentation outputs inherently support co-discovery by enabling unsupervised grouping and exploration of novel or anomalous samples in large datasets. By leveraging model-driven boundaries, researchers can more efficiently group and investigate novel or unusual samples, thereby streamlining the identification of interesting features and patterns within large datasets. This process is further enhanced by jointly analyzing embedding spaces relative to these segmentations, which provides deeper insight into ecological patterns and variability. To accelerate exploration, models can be deployed for collaborative, open-ended discovery, allowing us to rapidly sift through extensive data volumes and surface new, atypical, or noteworthy observations for further investigation \cite{Zhang2023, Lu2024}.

\section{Open Research Section}
\label{sec9}

\subsection{Materials and Tools}
\label{subsec24}

The experimentation was done in Python 3.9.13. To meet the system’s scientific objectives and maximize reliability, SIT-FUSE relies on a broad ecosystem of mature, well-supported open-source packages for geospatial analytics, large-scale data processing, and machine learning \cite{LaHaye2025c}. The architecture incorporates numpy, scipy, dask, xarray, Zarr, numba, and cupy for efficient management and computation of large datasets on both CPUs and GPUs \cite{Lam2015, Hoyer2017, Harris2020, cupy_learningsys2017, Virtanen2020, zarr2025}. Sci-kit-learn, PyTorch, and torchvision handle machine learning model training, deployment, and automatic differentiation \cite{Pedregosa2011, Paszke2019}, while additional RBM-based functionality is enabled via the Learnergy library, as PyTorch does not natively support Restricted Boltzmann Machines \cite{Roder2020}.

For geospatial data workflows, key open-source tools include pyresample, GDAL, OSR, healpy, polar2grid, and GeoPandas \cite{geopandas, Hoese2025, Rouault2025}. Classical computer vision tasks leverage OpenCV \cite{opencv}. Combining these robust, widely adopted technologies enables rapid integration of advanced methods. It supports the modular, maintainable design of SIT-FUSE, with most of the underlying complexity abstracted from the end user. SIT-FUSE is maintained as a public repository on GitHub and continues to prioritize open access in its ongoing development. Context assignment and visualization tasks utilize QGIS, a widely used open-source GIS platform \cite{QGISsoftware}. For this work, SIT-FUSE models were deployed on a server with a NVIDIA GeForce Titan V100 GPU equipped with 32 GB of memory.

\subsection{Produced Data, Models, and Code}
All data supporting the findings of this study are openly available on Zenodo (\url{https://doi.org/10.5281/zenodo.15693706}). The trained model weights and configuration files can be accessed via HuggingFace (\url{https://doi.org/10.57967/HF/5837}). The complete software codebase is publicly archived and available on GitHub at the time of submission (\url{https://zenodo.org/records/17117149}); \cite{LaHaye2025c, LaHaye2025d, LaHaye2025e}.

\section{Conflicts of Interest Declaration}
\label{sec8}
The funding agencies had no involvement in the study’s design, data collection, analysis, or interpretation, nor in the writing of the manuscript or the decision to publish these results. The authors declare there are no conflicts of interest for this manuscript.

Other affiliations held by the authors during this study include UCLA / JIFRESSE, Chapman University, and MLAT Lab.

\section{Funding}
\label{sec7}
This research received financial support from the Jet Propulsion Laboratory’s Office of Research and Development. Computational analyses were conducted using resources provided by both the Jet Propulsion Laboratory and the MLAT Laboratory at the Fowler School of Engineering, Chapman University.

\acknowledgments
The authors thank NASA, the Jet Propulsion Laboratory (JPL), Caltech, the Machine Learning and Affiliated Technologies (MLAT) Laboratory, the Spatial Informatics Group, LLC, and the Schmid College of Science and Technology at Chapman University for their crucial support and contributions to this research. We extend our thanks to Christian Frankenberg and his laboratory at Caltech, as well as NASA, for providing critical datasets that enabled this work. \add[]{We thank the developers and maintainers of the Coastal Harmful Algal Bloom Risk Mapping (C‑HARM) system for providing model output and documentation that made the comparison analyses in this study possible. We are also grateful to the many scientists, technicians, and program managers who collect, curate, and distribute the in situ HAB and phytoplankton observations used here, including contributors to the California Harmful Algal Bloom Monitoring and Alert Program (CalHABMAP) and the Florida Fish and Wildlife Conservation Commission’s HAB monitoring programs, as well as the agencies that support these efforts. Their sustained investments in field sampling and data stewardship provide the foundation for evaluating and improving satellite‑based HAB monitoring frameworks such as SIT‑FUSE.} Finally, we appreciate the insightful comments and suggestions provided by the anonymous reviewers, which substantially improved the quality of this manuscript.

\bibliography{agusample}

\end{document}